\documentclass[lettersize,journal]{IEEEtran}
\usepackage{amsmath,amsfonts}
\usepackage{algorithm}
\usepackage{array}
\usepackage{textcomp}
\usepackage{stfloats}
\usepackage{url}
\usepackage{verbatim}
\usepackage{graphicx}
\usepackage{wrapfig}
\usepackage{tcolorbox}
\usepackage{amssymb}
\usepackage{threeparttable}
\usepackage{booktabs}
\usepackage{cite}

\usepackage{setspace}
\usepackage[utf8]{inputenc} 
\usepackage[T1]{fontenc}    
\usepackage{hyperref}       
\usepackage{url}            
\usepackage{booktabs}       
\usepackage{amsfonts}       
\usepackage{nicefrac}       
\usepackage{microtype}      
\usepackage{xcolor}         
\usepackage{wrapfig}
\usepackage{threeparttable}

\usepackage{amsmath,amssymb,amsfonts}
\usepackage{amssymb}
\usepackage{breqn}
\usepackage{ragged2e}
\usepackage{cleveref}
\usepackage{float}   
	\usepackage{subfigure}
\usepackage{booktabs}
\usepackage{cases}
\usepackage{graphicx}
\usepackage{multirow}
\usepackage{mathrsfs}
\usepackage{textcomp}
\usepackage{xcolor}
\usepackage{bm}
\usepackage[noend]{algpseudocode}
\usepackage{tcolorbox}
\usepackage{algorithmicx,algorithm}

\newtheorem{theorem}{Theorem}

\newtheorem{definition}{Definition}

\begin{document}

\title{Explainable Molecular Property Prediction: Aligning Chemical Concepts with Predictions via Language Models}

\author{
Zhenzhong Wang\textsuperscript{†},~\IEEEmembership{Member, IEEE},
Zehui Lin\textsuperscript{†}, 
Wanyu Lin$^{*}$,~\IEEEmembership{Member, IEEE},
Ming Yang,
Minggang Zeng,\\
Kay Chen Tan,~\IEEEmembership{Fellow, IEEE}
\IEEEcompsocitemizethanks{\IEEEcompsocthanksitem
Zhenzhong Wang is with the Department of Computing, The Hong Kong Polytechnic University, Hong Kong SAR, P.R. China and Xiamen University, P.R. China (e-mail: zhenzhongwang@xmu.edu.cn).

Zehui Lin is with the Department of Data Science and Artificial Intelligence, The Hong Kong Polytechnic University, Hong Kong SAR, P.R. China (e-mail: zehui.lin@polyu.edu.hk).

Wanyu Lin is with the Department of Computing and Department of Data Science and Artificial Intelligence, The Hong Kong Polytechnic University, Hong Kong SAR, P.R. China (e-mail: wan-yu.lin@polyu.edu.hk).

Ming Yang is with the Department of Applied Physics, The Hong Kong Polytechnic University, Hong Kong SAR, P.R. China (e-mail: mingyang@polyu.edu.hk).

Minggang Zeng is with the Institute of High Performance Computing, Agency for Science, Technology and Research (A*STAR), Fusionopolis Way, \#16-16 Connexis, Singapore 138632, Republic of Singapore (e-mail: zengmg@ihpc.a-star.edu.sg).
    
Kay Chen Tan is with the Department of Data Science and Artificial Intelligence, The Hong Kong Polytechnic University, Hong Kong SAR, P.R. China (e-mail: kctan@polyu.edu.hk).

This work was supported in part by the National Natural Science Foundation of China under Grant No. U21A20512; in part by the Research Grants Council of the Hong Kong SAR under Grant C5052-23G, Grant PolyU 11211521, Grant PolyU 15218622, and Grant PolyU 15215623; in part by the Hong Kong Research Grant Council General Research Fund under Grant PolyU 15208222; in part by the National Natural Science Foundation of China Young Scientist Fund under Grant PolyU A0040473 and Start-Up Fund for New Recruits under Grant PolyU A0046682; \textit{\textsuperscript{†}These authors contributed equally to this work.} \textit{(*Corresponding author: Wanyu Lin.)} 

	}	
}

\maketitle
\begin{abstract}

Providing explainable molecular property predictions is critical for many scientific domains, such as drug discovery and material science. Though transformer-based language models have shown great potential in accurate molecular property prediction, they neither provide chemically meaningful explanations nor faithfully reveal the molecular structure-property relationships. In this work, we develop a framework for {\em e}xplainable {\em mol}ecular property prediction based on {\em la}nguage models, dubbed as {\em Lamole}, which can provide chemical concepts-aligned explanations. We take a string-based molecular representation --- Group SELFIES --- as input tokens to pre-train and fine-tune our {\em Lamole}, as it provides chemically meaningful semantics. By disentangling the information flows of {\em Lamole},  we propose considering both self-attention weights and gradients for better quantification of each chemically meaningful substructure's impact on the model's output. To make the explanations more faithful to the structure-property relationship, we then carefully craft a marginal loss to explicitly optimize the explanations to align with the chemists' annotations. We bridge the manifold hypothesis with the elaborated marginal loss to prove that the loss can align the explanations with the tangent space of the data manifold, leading to concept-aligned explanations. Experimental results over eight datasets demonstrate {\em Lamole} can achieve comparable prediction accuracy and boost the explanation accuracy by up to $14.3\%$, being the state-of-the-art in explainable molecular property prediction. To further illustrate the actionable utility of the explanations derived from {\em Lamole}, we integrated the framework with an evolutionary algorithm. This integration established an interpretable optimization pipeline for molecular editing, demonstrating that {\em Lamole} functions beyond simple post-hoc analysis but serves as a practical guide for molecule discovery. 

\end{abstract}

\begin{IEEEkeywords}
explainability, molecular property prediction, language models.
\end{IEEEkeywords}

\section{Introduction}
\label{submission}
\IEEEPARstart{M}{olecular} property prediction aims to reveal the structures-property relationships, assisting scientists in screening molecules for various applications such as drug discovery and material design~\cite{wang2024crystalline,fang2022geometry,deng2023systematic,NEURIPS2023_6a69d44b,NEURIPS2023_51ec452c,hong2024diffusion,kurban2022rapidly,kurban2025accelerating,kurban2026multimodal,kurban2021rare}. 
Several learning-based models are devised based on the underlying molecular representations, such as graph-based and string-based molecular representations. Among them, string-based molecular representations, e.g., simplified molecular input line entry systems (SMILES~\cite{weininger1988smiles}), stand out for their simplicity and adaptability~\cite{deng2023systematic,wigh2022review,cheng2022group}. By viewing the string-based molecular representation as a form of "chemical" language, the Transformer-based language models (LMs) like Bert~\cite{kenton2019bert} offer higher throughput and accuracy for molecular property prediction~\cite{deng2023systematic,ChemBERTa}.

  \begin{figure*}[!t] 
  \centering   
  \includegraphics[width=14.2cm]{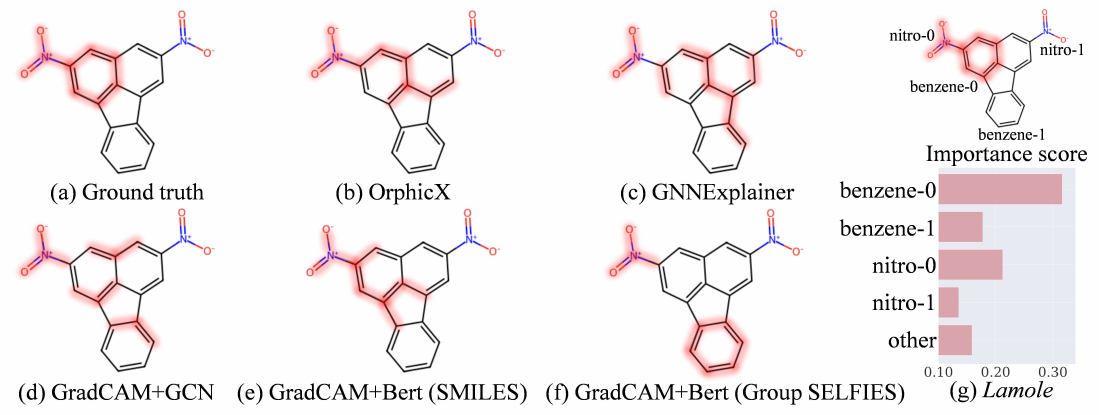}
  \caption{
  (a) The molecule visualization of prediction/explanation. The interaction between the benzene ring and the nitro group (highlighted in red) induces the mutagenic property of the molecule. (b)-(e) are the explanation results obtained with various methods: (b) OrphicX~\cite{lin2022orphicx}; (c) GNNExplainer~\cite{gnnexplainer}, (d) GNN with gradient-based explainability technique (GradCAM~\cite{Selvaraju_2017_ICCV}); (e) Bert with GradCAM (molecular string SMILES as input); (f) Bert with GradCAM (molecular string Group SELFIES~\cite{cheng2022group} as the input representation); (g) Our method \emph{Lamole} assigns an importance score to each functional group/fragment to indicate their contribution to the property.}
  \label{fig:intro}
\end{figure*}

Despite the superior performance of learning-based prediction methods, {\em what key factors induce the model's predictions remain largely unexplored}, impeding further advancements in the scientific domains. Typically, it is crucial to obtain explanations of predictions while achieving accurate predictions. These obtained explanations could be used for scientific hypotheses validation or/and providing actionable insights for refining investigations, such as optimization for molecular structural design~\cite{wu2023black,wellawatte2023perspective,das2022crysxpp}. With different types of molecular representations, explainability techniques for graph neural networks (GNNs) or LMs might be adopted to alleviate the general lack of explainability in molecular prediction~\cite{proietti2024explainable,gnnexplainer,NEURIPS2023_14cdc901,lin2021generative}.

However, we argue that existing explainability techniques often struggle to generate plausible explanations that can highlight chemically meaningful substructures and faithfully uncover the structure-property relationships simultaneously. Specifically, 1) from the molecular representation perspective, the commonly used representations do not explicitly encode the chemically meaningful substructures; current explainability methods can only highlight individual atoms and bonds as explanations (see Fig.~\ref{fig:intro} (b)$\sim$(e)). 2) From the perspective of explainability techniques, current methods suffer from two main limitations. First, they cannot effectively capture the interactions between functional groups within the molecular structure. Second, they could not generate explanations that align with chemists' intuition. As a result, they fail to produce explanations that faithfully reflect the structure-property relationships.
(see Fig.~\ref{fig:intro} (f)). Consequently, they cannot directly provide actionable utility to guide the rational design or editing of molecules to optimize desired properties. Therefore, an effective framework is imperative for explainable molecule property predictions.

\begin{figure}[!t]
\centering
\includegraphics[width=4.5cm]{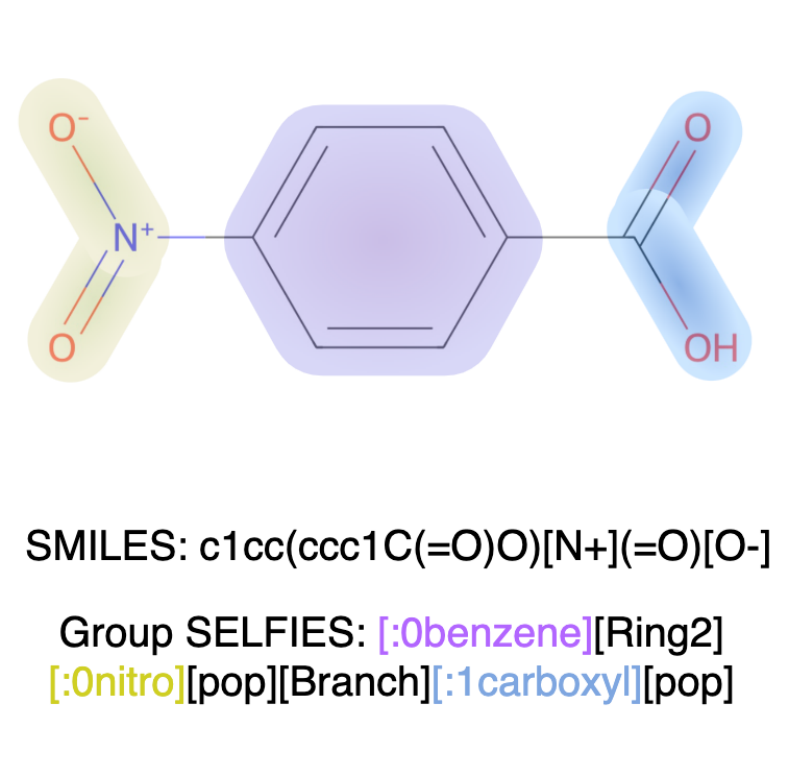}
\caption{The SMILES and Group SELFIES strings of p-nitrobenzoic acid molecule ($\mathrm{C_7H_5NO_4}$): The tokens in the Group SELFIES string highlighted by color are the corresponding functional groups.}
\label{fig:SELFIES}
\end{figure}

The recently proposed string-based molecular representation --- Group SELFIES~\cite{cheng2022group} --- encodes molecules at the functional group/fragment level, showcasing the possibility of obtaining chemically meaningful explanations. As shown in Fig. \ref{fig:SELFIES}, Group SELFIES converts a p-nitrobenzoic acid molecule to a string, which explicitly encodes chemically meaningful substructures as tokens, including a benzene, a nitro group, and a carboxyl group. Compared with 2D molecular graphs, Group SELFIES provides inherent semantic information, making it easier for the model to capture and understand chemically meaningful semantics. Moreover, using Group SELFIES eliminates the need to identify or segment chemically meaningful substructures in 2D molecular graphs.
With Group SELFIES's simplicity and adaptability, this work develops an {\em e}xplainable {\em mol}ecular property prediction framework based on {\em la}nguage models to provide chemical concepts-aligned explanations, called {\em Lamole} (see Fig.~\ref{fig:intro} (g)). The contributions can be summarized as follows.

\begin{enumerate}

 \item We found that existing explainability techniques fail to reveal molecular structure-property relationships faithfully. Therefore, to provide explanations directly aligned with chemical intuition, we first leverage the Group SELFIES representation—which inherently encodes chemically meaningful substructures—for both pre-training and fine-tuning our LMs. By integrating this chemically grounded semantic input, our framework enables LMs to directly comprehend and operate on functional group semantics, moving beyond low-level features to generate explanations that are immediately relevant and intuitive for chemists.

 \item We analyzed the limitations of existing methods in revealing molecular structure-property relationships, noting that approaches relying solely on attention or gradients as explanations often fail to adequately reflect the model's underlying reasoning process. For an explanation to be faithful, the generation process must be closely coupled with the mechanism of the model architecture. Therefore, by carefully disentangling the information flows within Transformer-based LMs, we propose an information flow-based explainable method. This strategy is designed specifically to capture the interactions between chemically meaningful substructures, thereby providing a more accurate quantification of each substructure's contribution to the predicted molecular properties.
 

 
    \item To make the explanations more faithful to the structure-property relationships, we elaborate on one marginal loss to calibrate the explanations by aligning them with the chemists' annotations. Empirically, we show that using only a few molecules with ground truth annotations can significantly improve the explanation accuracy by up to 5\%. Furthermore, theoretically, we first bridge the manifold hypothesis with explainable molecular property prediction. We theoretically demonstrate that the elaborated marginal loss aligns explanations with the data manifold, respecting the structure-property relationship.

   \item We further demonstrate the actionable utility of the explanations derived from {\em Lamole} through a dedicated molecular editing case study. By integrating  {\em Lamole}'s identified key functional groups with an evolutionary algorithm, we establish an interpretable optimization pipeline that guides the molecular optimization process to enhance target properties. This integration showcases {\em Lamole}'s capability to bridge predictive insights with rational molecular design and optimization, paving the way for more efficient chemical discovery.

\item We evaluated our method across eight diverse datasets spanning both classification and regression tasks, including a wide range of molecular properties such as hepatotoxicity, mutagenicity, water solubility, and permeability. Experimental results demonstrate that our proposed {\em Lamole} achieves comparable predictive accuracy relative to other established baselines while exhibiting superior explanation accuracy, confirming the broad applicability of {\em Lamole} across different property prediction tasks.

\end{enumerate}

\section{Related Work}

Molecular property prediction is a fundamental task aimed at elucidating the relationship between molecular structure and function. Various learning-based models have been developed based on distinct molecular representations, primarily categorized into graph-based and string-based formats. $\text{GNN}$s explicitly model molecular systems by incorporating atomic positions and pairwise interactions through message passing. For instance, $\text{SchNOrb}$ \cite{schutt2019unifying} adapted distance-based convolution by integrating symmetry-adapted atomic orbital features for electronic structure prediction. $\text{MiCaM}$ \cite{geng2023de} introduced a fragment-based deep generative model that enhances novelty by automatically discovering connection-aware motifs through iterative subgraph merging, which are then utilized in a dedicated generator for de novo molecular generation. Alternatively, string-based molecular representations, such as $\text{SMILES}$, are favored for their simplicity and adaptability within sequential models. $\text{TamGen}$ \cite{wu2024tamgen}, for example, leveraged a $\text{GPT}$-like chemical language model to achieve target-aware molecule generation and compound refinement. This work demonstrated the efficacy of $\text{GPT}$-like architectures and training strategies for generation within the one-dimensional $\text{SMILES}$ chemical space by integrating specific modules for encoding protein structure and seeding compound information.

Several explainable GNNs are proposed to explain the relationship between the input graph and the prediction~\cite{wang2024generating,lin2022orphicx,lin2021generative,sun2023explainable,xiong2019pushing,gnnexplainer,pgexplainer}. Among these works, structure similarity or attention weights are proposed to capture structural interaction~\cite{sun2023explainable,xiong2019pushing}. However, two similar substructures do not necessarily lead to interaction between them, and attention weights are often inconsistent with the feature  importance~\cite{jain2019attention,serrano2019attention,abnar2020quantifying}. In addition, as shown in Fig.~\ref{fig:intro} (b)$\sim$(d), some trivial structures received relatively high importance scores, indicating the explanations might not align well with the chemical concepts.


On the other hand, with string-based molecular representations, LMs show great potential in molecular property prediction~\cite{ChemBERTa,Smilesbert,chemberta2,ross2022large}. However, the "black-box" characteristics of LMs hamper trust use of these potent computational tools in scientific domains. Some explainability techniques could be applied to LMs. One way is to use the attention weights over the input tokens. However, recent studies suggest that "attention is not explanation" because attention weights could not reflect the true feature  importance~\cite{jain2019attention,serrano2019attention,abnar2020quantifying}. Perturbation-based methods perturb the inputs and evaluate the output changes to reveal the input importance. However, the generated explanations may change drastically with very small perturbations~\cite{agarwal2021towards}. 
Gradient-based methods determine the feature importance by the partial derivatives of the output to each feature~\cite{Selvaraju_2017_ICCV}. However, several works show that the gradient-based methods may not be reliable, as they disregard the influence of model architectures on the output and fail to incorporate the information of the model architectures into the explanations~\cite{NEURIPS2018_294a8ed2,agarwal2021towards,rudin2019stop}. Therefore, the explanation generation process should reflect the model reasoning process behind the model architectures.
To this end, we disentangle the model architectures' information flows to generate explanations that faithfully reveal the structure-property relationship.


\section{Methodology}

\textbf{Problem Setup.} Given a dataset $\mathcal{G}=\{(g^{(i)},y^{(i)})\}$ consisting of molecular graphs $\{g^{(i)}\}$ with their property labels $\{y^{(i)}\}$, explainable molecular property prediction aims to train a model $f$ to map a molecule $g$ to its property $y$, denoted as denoted as $f: g \mapsto y$, while providing an importance score vector $\mathbf{v}^{(i)}=\{\mathbf{v}_1^{(i)},\ldots,\mathbf{v}_j^{(i)}\}$ to indicate the contribution of $j$-th functional group/fragment to its property $y$.
Particularly, this work proposes to convert the molecular graph $g^{(i)}$ into the Group SELFIES string represented as $s^{(i)}=\{t_1^{(i)},\ldots,t_j^{(i)}\}$, where $t_j^{(i)}$ is the $j$-th functional group/fragment's token, i.e., [$\cdot$] in Fig.~\ref{fig:SELFIES}. In addition, this work calibrates the explanations in a supervised manner. For this purpose, a few molecules with annotation masks are provided. The annotation masks $\mathbf{m}(g^{(i)})\in\{0,1\}^j$ indicate whether $t_j^{(i)}$ is the token of a ground truth substructure, where $\mathbf{m}_j(g^{(i)})=1$ denotes the substructure corresponding to the token $t_j^{(i)}$ inducing the molecular property of $g^{(i)}$.
This work uses $\mathcal{D}=\{(s^{(i)},y^{(i)})\}$ with a few annotation masks to learn a model $f$ for explainable molecular property prediction. We omit the superscript $^{(i)}$ for simplicity in the following parts.

\begin{figure*}[t] 
  \centering   
  \includegraphics[width=15.2cm]{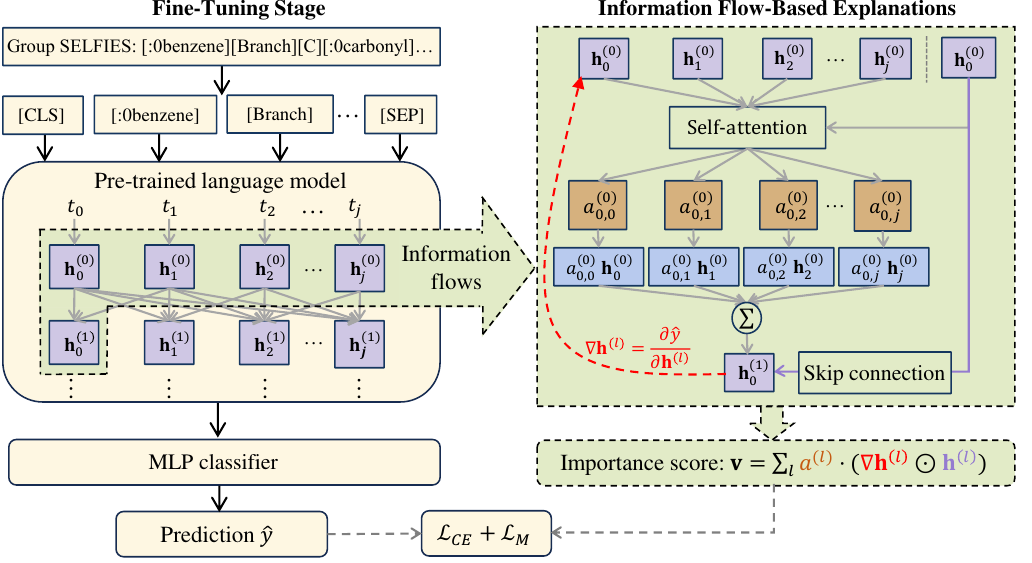}
  \caption{An illustration of \emph{Lamole}. Left panel:  Group SELFIES strings are tokenized for fine-tuning the pre-trained language model, and an MLP classifier is equipped with a cross-entropy loss $\mathcal{L}_{CE}$ for molecular property prediction. Right panel:
  We disentangle the information flows of the Transformer to assert that both attention weights and gradient determine the output. Therefore, we incorporate the attention weights and gradients together to generate importance scores $\mathbf{v}$ as explanations. In addition, a marginal loss $\mathcal{L}_{M}$ is designed to align explanations with the chemists’ annotations $\mathbf{m}$.}
  \label{fig:icmlframework}
\end{figure*}

\subsection{Our Design: Lamole}
In this work, we pre-train Transformer-based LMs, e.g., Bert family models, using the Group SELFIES corpus to make the models understand the chemical semantics behind Group SELFIES strings.
Then, we fine-tune the LMs with Group SELFIES strings and the molecular property labels for explainable molecular property prediction.  An illustration of proposed \emph{Lamole} is in Fig.~\ref{fig:icmlframework}. In what follows, we will introduce the detailed design of~\emph{Lamole}.

\textbf{Fine-Tuning Stage.} Fig. \ref{fig:icmlframework} shows the fine-tuning stage of the proposed \emph{Lamole}. We assume that the Transformer encoder in \emph{Lamole} stacks $L$ identical Transformer layers to encode the molecular string $s$ as token embedding $\mathbf{h}^{(l)}=\{\mathbf{h}_1^{(l)},\ldots,\mathbf{h}_j^{(l)}\}$, where $\mathbf{h}^{(l)}$ is the token embedding at the $l$-th layer.
We use the self-attention weighted average embedding $\mathbf{h}_o=(\sum_{j=1}\alpha_j\cdot\mathbf{h}_j^{(L)})/(\sum_{j=1}\alpha_j)$ for molecular property prediction, where $\alpha_j$ is the attention weight of the $j$-th token.
A multilayer perceptron (MLP) classifier is added to predict the molecular property $\hat{y}=\mathrm{MLP}(\mathbf{h}_{o})$ by minimizing the classification cross-entropy loss $\mathcal{L}_{CE}(y,\hat{y})$.

{\textbf{Analysis of Foundational Explanation Methods.}
Before introducing our information flow-based approach, we first analyze two foundational methods—attention and gradients—and detail their limitations when applied to Transformer-based LMs.

1) Attention as Explanations: This approach quantifies the importance of each molecule token to the $\text{LM}$'s final decision. This is achieved by extracting the softmax cross-attention from the starting answer token to all input tokens across all layers of the backbone LM. The resulting prediction-to-token attention is represented as $A_{\text{st}}(s) \in \mathbb{R}^{L \times H \times 1 \times T}$, where $L$ and $H$ denote the number of layers and heads-per-layer, respectively, and $T$ is the total number of tokens provided. We subsequently compute the prediction-to-token attention $\hat{A}_{\text{st}}(s)$ by averaging across attention heads: $\hat{A}_{\text{st}}(s) = \frac{1}{H} \sum_{h=1}^H A_{\text{st}}(s)$. However, recent studies suggest that ``attention is not explanation" because the attention scores often fail to reflect the true importance of the input features to the model's decision~\cite{jain2019attention,serrano2019attention,abnar2020quantifying}.

2). Gradients as Explanations: This common practice utilizes gradient-based methods to determine feature importance by analyzing the partial derivatives of the output with respect to each input feature \cite{Selvaraju_2017_ICCV}. We seek to identify relevant tokens by computing the gradient of the LM's decision with respect to the input tokens. The pure gradient is versatile as it is not restricted to a Transformer architecture. Specifically, for a molecule-label pair $(s, y)$, the importance is computed as $G(s, y) = \|\nabla_s v(s, y)\|_2$, where $v(s, y)$ is the logarithm of the maximum output probability of the LM.

While both practices are widely used to derive explanations, we argue that relying on gradients alone is insufficient for effectively capturing substructure interactions. As illustrated in Fig.~\ref{fig:intro} (f), a pure gradient-based method like GradCAM can inaccurately attribute a property to an unrelated structural fragment, such as incorrectly highlighting a distant benzene ring alongside the true nitro group contributor. The primary reason for this deficiency lies in the internal architecture of the Transformer. As shown in the right panel of Fig. \ref{fig:icmlframework}, the output is influenced by a confluence of factors, including attention weights, gradients, and the original input itself, primarily due to the ubiquitous skip connections and the attention mechanism. Consequently, using gradients in isolation fails to capture the rich network of information exchange—the interactions—that occurs between substructures via the attention layers. To address this limitation, we leverage both attention and gradients, integrating them alongside the input to derive a more comprehensive explanation. The following section details this integration of attention weights into our gradient-based explanation framework.

\textbf{Information Flow-Based Explanations.} 
Firstly, we show the process of deriving the gradient-based explanations. Similar to GradCAM~\cite{Selvaraju_2017_ICCV}, the gradient with respect to the $j$-th token's embedding $\mathbf{h}_j^{(l)}$ at the $l$-th layer is derived by $\nabla \mathbf{h}_j^{(l)}=\partial \hat{y}/{\partial \mathbf{h}_j^{(l)}}$,
where $\nabla \mathbf{h}_j$ signifies the importance of the $j$-th token in relation to the predicted property $\hat{y}$. Due to the skip connection in Fig. \ref{fig:icmlframework} right panel, the input, and its corresponding gradient should be leveraged together, and the weighted importance $\mathbf{w}$ of the $j$-th token at the $l$-th layer can be determined by
\begin{equation}
    \mathbf{w}_{j}^{(l)}= \nabla \mathbf{h}_j^{(l)} \odot \mathbf{h}_j^{(l)},
\end{equation}
where $\odot$ is the Hadamard product. The weighted importance is regarded as the gradient-based explanation.

The interaction among tokens can be revealed by the self-attention mechanism in Fig. \ref{fig:icmlframework} right panel. The attention mechanism calculates pairwise similarity scores between all pairs of tokens to determine attention weights, and these attention weights inherently encode the functional group interactions.
Therefore, we combine the attention weights with the gradient-based explanation to capture functional group interactions.   Assuming the attention weights of the $j$-th token at the $l$-th layer is $\alpha_j^{(l)}$, we integrate the attention weights with weighted importance $\mathbf{w}$ to consider the interactions. The importance score of the molecule $g$ can be obtained by
\begin{equation}
    \mathbf{v}_j^{(l)}(g)= \left(\tanh(\overline{\alpha_j^{(l)}})\cdot\tanh(\mathbf{w}_j^{(l)})\right)^{\frac{1}{2}},
\end{equation}
where $\overline{\alpha_j^{(l)}}$ is the averaging of attention weights of multiple attention heads. 
Finally, we sum $\mathbf{v}_i^{(l)}$ over all layers as the final importance score of the $j$-th token, 
\begin{equation}
    \mathbf{v}(g)= \mathrm{softmax}(\sum_{l=1}^{L}\mathbf{v}^{(l)}(g)).
\end{equation}
The higher the importance scores, the greater the contribution of the corresponding functional groups/fragments to the molecular property.

\textbf{Towards Plausible Explanations.} One plausible explanation should faithfully uncover the structure-property relationships. In other words, the explanation should match the ground-truth substructures with high confidence. Nevertheless, the importance scores of ground-truth substructures might not be significantly higher than those of other parts. To address this issue, we propose a marginal loss to explicitly align explanations with the chemists' annotations to improve the explanations' plausibility.

First, we formally define the plausibility of explanations. "Plausibility" refers to how the interpretation convinces humans~\cite{wiegreffe2019attention,herman2017promise,jacovi2020towards}. Similarly, in our context, "plausibility" refers to the degree of confidence in the explanations that would convince the chemists.

Eq. (\ref{eq:epmetric}) defines the explanation plausibility based on the scores of two parts, i.e., the scores on ground truth and the scores on non-ground truth. The lower the scores on non-ground truth and the greater the scores on ground truth, the better the explanation plausibility. Therefore, to maximize the plausibility, our objective can be transformed to minimize the importance score of non-ground truth and maximize the importance score of ground truth.

\begin{tcolorbox} [boxrule=.0mm,arc = 0mm, outer arc = 0mm,left=0.2mm,right=0.2mm,top=0.2mm,bottom=0.2mm]
\begin{definition}
    (Plausibility): Given the importance scores $\mathbf{v}$ over all tokens in the molecule $g$, the mean importance score $\mathbf{v}_{\in \mathcal{T}_g}$ over ground truth substructures $\mathcal{T}_g$ and the mean importance score $\mathbf{v}_{\notin \mathcal{T}_g}$ over other substructures $\mathcal{T}_g$ are denoted by $\mathbf{v}_{\in \mathcal{T}_g}=\frac{\sum_j \mathbf{v}_j \cdot\mathbb{I}(t_j\in \mathcal{T}_g)}{\sum_j \mathbb{I}(t_j\in \mathcal{T}_g)}$ and $\mathbf{v}_{\notin \mathcal{T}_g}=\frac{\sum_j \mathbf{v}_j \cdot\mathbb{I}(t_j\notin \mathcal{T}_g)}{\sum_j \mathbb{I}(t_j\notin \mathcal{T}_g)}$, respectively, where $\mathbb{I}(\cdot)$ is the indicator function. The explanations' plausibility $\mathrm{EP}(g)$ is defined as the ratio of the difference between $\mathbf{v}_{\in \mathcal{T}_g}$ and $\mathbf{v}_{\notin \mathcal{T}_g}$ to $\mathbf{v}_{\notin \mathcal{T}_g}$,
        \begin{equation}\label{eq:epmetric}
\mathrm{EP}(g)=\frac{\mathbf{v}_{\in \mathcal{T}_g}-\mathbf{v}_{\notin \mathcal{T}_g}}{\mathbf{v}_{\notin \mathcal{T}_g}}.
\end{equation}
\end{definition}
The higher the $\mathrm{EP}$ value, the greater the confidence of the explanation in matching the ground truth substructure.
\end{tcolorbox}

To this end, we design a max-margin loss to optimize the importance score. In our work, the ground truth substructures are annotated by a binary mask vector $\mathbf{m}(g)\in\{0,1\}^{j}$. It is worth noting that using only a few annotations can significantly improve the explanation accuracy. Specifically, the mask vector $\mathbf{m}$ enforces the explanations to align with the ground truth substructures. To achieve the goal, a max-margin loss is designed by maximizing the mean value of the importance scores of tokens that have mask values of 1 while minimizing the mean value of importance scores for tokens with mask values of 0.

\begin{multline}\label{eq:l1}
    \mathcal{L}_{M}(\mathbf{v},\mathbf{m}) = \mathbb{E}_{g\in\mathcal{G}} \Bigg[ \max\Bigg( 0, \frac{\sum_{j=1}(1-\mathbf{m}_j(g))\cdot \mathbf{v}_j(g)}{N_s} \\
    - \frac{\sum_{j=1} \mathbf{m}_j(g)\cdot \mathbf{v}_j(g)}{N_c} \Bigg) + \triangle_1 \Bigg],
\end{multline}
where $\triangle_1$ is a margin term, $N_s$ is the number of tokens with mask values $\mathbf{m}(g)$ of 0, and $N_c$ is the number of tokens with mask values $\mathbf{m}(g)$ of 1.
The overall optimization objective of the fine-tuning stage is to minimize $\mathcal{L}_{CE}+\mathcal{L}_{M}$. The core of Eq. (\ref{eq:l1}) is the discrepancy between the average importance score of ground truth and the average importance score of non-ground truth. By minimizing $\mathcal{L}_{M}$, the discrepancy between the two importance scores is maximized. In other words, the average importance score of non-ground truth is suppressed, and the average importance score of ground truth is increased. Finally, the explanation plausibility defined in Eq. (\ref{eq:epmetric}) is improved. The next section will theoretically show that by using the designed marginal loss, the explanations can faithfully reflect the structure-property relationships.



\subsection{Theoretical Analysis}
We bridge the manifold hypothesis with the marginal loss to theoretically show that the explanations can respect the structure-property relationships. 
Before giving the proof, the notation and definition regarding the manifold hypothesis are presented.

\textbf{Manifold Hypothesis.} It is widely believed that natural data, including molecules, distribute around a manifold~\cite{bordt2023manifold,lin2022orphicx,godwin2022simple,singh2020chemoverse}. 
According to the manifold hypothesis for gradient-based explanations~\cite{bordt2023manifold}, if a feature lies in the tangent space of a manifold, then the feature respects the manifold and contributes to the class, and such a feature is desirable to be explained. We call these features "causal features" in our work. Conversely, if a feature is orthogonal to the manifold, then the feature does not contribute to the class. We call these features "spurious features".

With the annotation masks, the causal features $s^*$ and spurious features $\overline{s}^{*}$ can be distinguished by $s^*=s\odot\mathbf{m}(g)$ and $\overline{s}^{*}=s\odot (1-\mathbf{m}(g))$, respectively, where $\overline{s}^*\cup s^*=s$ and $s^* \cap \overline{s}=\varnothing$. By projecting the causal features and spurious features into the data manifold $\mathcal{M}$, the corresponding manifold regarding the causal features and spurious features can be defined as follows,

\begin{tcolorbox} [boxrule=.0mm,arc = 0mm, outer arc = 0mm,left=0.3mm,right=0.3mm,top=0.3mm,bottom=0.3mm]
\begin{definition}
 (Causal feature manifold and spurious feature manifold): Assume the distribution $p(g|y)$ is implicitly modeled by a manifold $\mathcal{M}$, and the manifold can be decomposed into two components,
\begin{equation}\label{eq:decom}
\underbrace{ p(g|y)}_{\mathcal{M}}= \underbrace{p(g|y) \odot \mathbf{m}(g)}_{\mathcal{M}_c}+ \underbrace{p(g|y) \odot (1-\mathbf{m}(g))}_{\mathcal{M}_s},
\end{equation}
where $\mathcal{M}_c$ is the causal feature manifold and $\mathcal{M}_s$ represents the spurious feature manifold.
\end{definition}
\end{tcolorbox}

With this decomposition, we demonstrate how the gradient-based explanations $\nabla_g \log p(y|g)$ can uncover the structure-property relationships. 


\begin{tcolorbox} [boxrule=.0mm,arc = 0mm, outer arc = 0mm,left=0.4mm,right=0.4mm,top=0.4mm,bottom=0.4mm]
\begin{theorem} 
    The marginal loss of Eq. (\ref{eq:l1}) aligns the gradient-based explanations $\nabla_g \log p(y|g)$ with the tangent space of the causal feature manifold $\mathcal{M}_c$, thus respecting the structure-property relationships.
\end{theorem}
\end{tcolorbox}

\textbf{Proof of Theorem 1}: The gradient with respect to the prediction $\nabla_g \log p(y|g)$ can be decomposed into the gradient on the causal features and spurious features, respectively,

\begin{multline}\label{eq:decom}
    \nabla_g \log p(y|g) = \nabla_g \log p(y|g) \odot \mathbf{m}(g) \\
    + \nabla_g \log p(y|g) \odot (1-\mathbf{m}(g)).
\end{multline}
By minimizing the loss of Eq. (\ref{eq:l1}), the gradients on spurious features $\nabla_g \log p(y|g) \odot (1-\mathbf{m}(g))$ are suppressed, and $\nabla_g \log p(y|g)$ approximates $\nabla_g \log p(y|g) \odot \mathbf{m}(g)$. Therefore, we have $\nabla_g \log p(y|g)\approx\nabla_g \log p(y|g) \odot \mathbf{m}(g)$. On the other hand, $\nabla_g \log p(y|g) \odot \mathbf{m}(g)$ can be rewritten as
\begin{multline}\label{eq:decomp2}
    \nabla_g \log p(y|g) \odot \mathbf{m}(g) = \nabla_g \log p(g|y) \odot \mathbf{m}(g) \\
    - \sum_{j} p(y=j|g) \nabla_g \log p(g|y=j) \odot \mathbf{m}(g).
\end{multline}
Because the data distribution $p(g|y)\odot \mathbf{m}(g)$ reflects the causal feature manifold $\mathcal{M}_c$, the gradient of the distribution $\nabla_g p(g|y)\odot \mathbf{m}(g)$ represents the tangent space of the causal feature manifold $\mathcal{M}_c$. In addition, Eq. (\ref{eq:decomp2}) shows that the $\nabla_g \log p(y|g)\odot \mathbf{m}(g)$ is a linear combination of $\nabla_g p(g|y) \odot \mathbf{m}(g)$, so $\nabla_g \log p(y|g)\odot \mathbf{m}(g)$ also lies tangent space of the manifold $\mathcal{M}_c$. 

Together with Eq. (\ref{eq:decom}) and Eq. (\ref{eq:decomp2}), we prove the gradient-based explanations $\nabla_g \log p(y|g)$ lies tangent space of the manifold $\mathcal{M}_c$. This indicates by minimizing the loss of Eq. (\ref{eq:l1}), the model $p(y|g)$ has reflected the causal feature manifold.
According to the manifold hypothesis, the features on the causal feature manifold contribute to the molecular property. Therefore, the gradient-based explanations $\nabla_g \log p(y|g)$ can uncover the causal features, thus revealing the structure-property relationships. This completes the proof.

\section{Experiments}
\subsection{Experimental Setup}

\textbf{Datasets}. 
 We use six datasets on two types of tasks, i.e., hepatotoxicity and mutagenicity, to evaluate the algorithmic performance for the explainable molecular property prediction. Six mutagenicity datasets are Mutag~\cite{mutagdata}, Mutagen~\cite{tudataset}, PTC-FM~\cite{ptcdata}, PTC-FR~\cite{ptcdata}, PTC-MM~\cite{ptcdata}, and PTC-MR~\cite{ptcdata}. For hepatotoxicity~\cite{ptcdata}, the Liver dataset~\cite{liu2015data} is used. Larger-sized molecules typically include more complex structures. The datasets that we used contained relatively large molecules. The maximal number of atoms of Mutag, Mutagen, PTC-FM, PTC-FR, PTC-MM, PTC-MR, and Liver are 26, 417, 64, 64, 64, 64, and 157, respectively. 
 The details of the used dataset are provided in Table \ref{tab:ourdatasets}.

\setlength{\tabcolsep}{1.5pt}
 \begin{table}[t]\footnotesize
  \centering
  \caption{Statistiscal Information of the Datasets}
    \begin{threeparttable}
    \begin{tabular}{cccccccc}
    \toprule
    Datasets & Mutag & Mutagen & PTC-FM & PTC-FR & PTC-MM & PTC-MR & Liver \\
    \midrule
    Graphs & 188   & 4337  & 349   & 351   & 336   & 344   & 587 \\
    Classes & 2     & 2     & 2     & 2     & 2     & 2     & 3 \\
    Max nodes& 26& 417 &64&64&64&64&157\\
    Avg nodes  & 17.9  & 29    & 14.1  & 14.6  & 14    & 14.3  & 25.6 \\
    Avg edges & 19.8  & 30    & 14.5  & 15    & 14.3  & 14.7  & 27.4 \\
    Ground truth$^{*}$ & 120   & 724   & 58    & 49    & 51    & 61    & 187 \\
    \bottomrule
    \end{tabular}%
         \begin{tablenotes}
        \footnotesize
        \item $^{*}$ denotes the number of molecules with known ground truth substructures.
      \end{tablenotes}
  \end{threeparttable}
  \label{tab:ourdatasets}%
\end{table}%


Following OrphicX~\cite{lin2022orphicx}, on the Mutag, Mutagen, PTC-FM, PTC-FR, PTC-MM, and PTC-MR datasets, we only consider the explanations for the mutagenic class, because the molecules of the non-mutagenic class have no ground truth. Although some works used single $\mathrm{N}=\mathrm{N}$, $\mathrm{NO_2}$, or $\mathrm{NH_2}$ as ground truth, this is not reasonable, as 32\% of non-mutagenic graphs in Mutagen containing at least single $\mathrm{NO_2}$ or $\mathrm{NH_2}$. In fact, the ground truth for the mutagenic class is the benzene with a chemical group on it, such as $\mathrm{N}=\mathrm{N}$, $\mathrm{NO_2}$, and $\mathrm{NH_2}$~\cite{lin2021generative,lin2022orphicx,Quantitativemutagenicity}.

For the Liver dataset, the molecules of possible hepatotoxicity with ground truth substructures and hepatotoxicity with ground truth substructures are collected for explainable molecular property prediction. 
The twelve ground truth substructures of the Liver dataset are shown in Fig. \ref{fig:livergt}.

To assess the generalizability of our method's explainable prediction ability, we performed an extended evaluation on two critical absorption, distribution, metabolism, and excretion (ADME) properties: solubility and permeability. The evaluation was conducted using the ZINC dataset~\cite{zinc}. Lacking expert-annotated references for these complex endpoints, we employed an automated, descriptor-driven approach to generate the required ground-truth functional group labels. The annotation procedure for the ground truth substructures in the dataset is detailed in the supplemental material.

\begin{figure}[t] 
  \centering   
  \includegraphics[width=\linewidth]{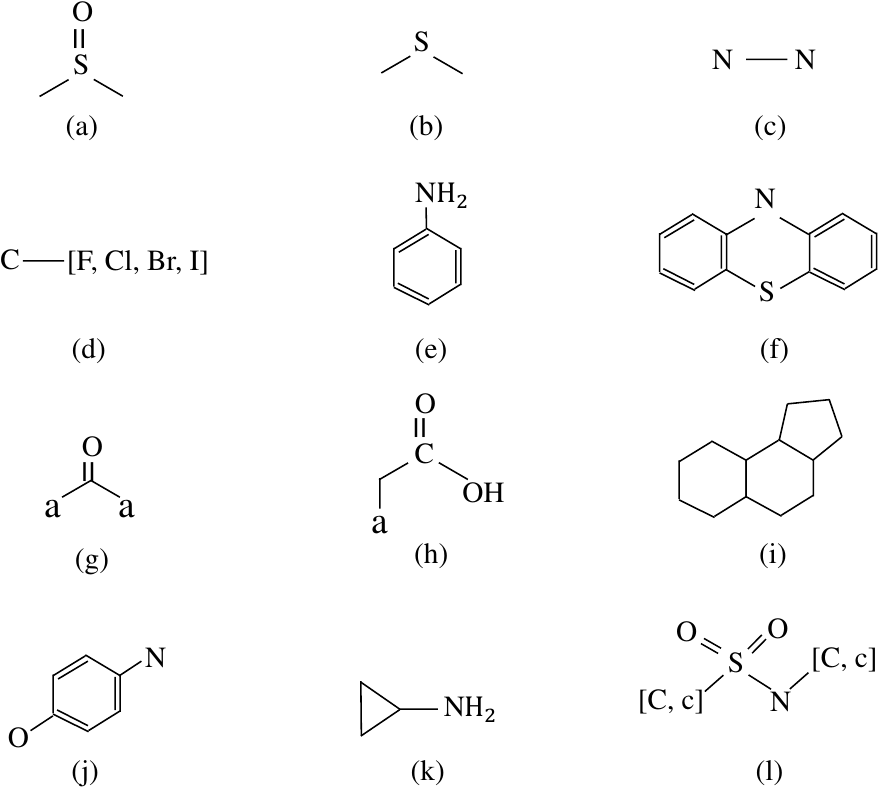}
  \caption{Twelve ground truth substructures of the Liver dataset. Lowercase element symbols represent aromatic atoms of the element; the letter "a" matches any aromatic atom. Elements in square brackets match any of the elements in a molecule.}
  \label{fig:livergt}
\end{figure}


\textbf{Evaluation Metrics.} We use four metrics for evaluating the performance of the proposed \emph{Lamole}. 
1) Classification Accuracy: We evaluate the model’s predictions by $\sum_{i=1}^{I}\mathbb{I}(y^{(i)}=\hat{y}^{(i)})/I$.
     2) Explanation Accuracy: We follow the experimental settings in GNNExplainer~\cite{gnnexplainer}, which formulates the explanation problem as a binary classification of edges. We treat edges inside ground-truth substructure as positive edges and negative otherwise, and AUC is adopted as the metric for quantitative evaluation. We only consider the mutagenic/hepatotoxic molecules because no explicit substructures exist in nonmutagenic/nonhepatotoxic ones. 
     3) Explanations' Fidelity: If an explanation is faithful, removing the most important fragments identified can cause a significant change in the model's prediction score for the target property. The definition is as follows: $\text{Fidelity}(G, G_{\text{ex}}) = [f(G)]_{c^*} - [f(G / G_{\text{ex}})]_{c^*}$, where $G$ is the original molecular inputs, $G_{\text{ex}}$ is the set of key functional groups/fragments identified, and $f(\cdot)$ is the model's classification probability or regression output.
     4) Explanations' Plausibility: We use the defined explanations' plausibility $\mathrm{EP}$ to measure how confident the explanation aligns with the ground truth.

\textbf{Baselines.} We combine \emph{Lamole} into three BERT family models, such as DistilBert~\cite{DistilBERT}, DeBerta~\cite{DEBERTA}, and Bert to evaluate the performance of the proposed \emph{Lamole}.
For evaluating classification accuracy, we compare our \emph{Lamole} with one SMILES string-based LM, ChemBERTa~\cite{ChemBERTa} and several GNNs including GCN~\cite{kipf2016semi}, DGCNN~\cite{zhang2018end},  edGNN~\cite{jaume2019edgnn}, GIN~\cite{xu2018powerful}, RW-GNN~\cite{nikolentzos2020random}, DropGNN~\cite{papp2021dropgnn}, and IEGN~\cite{maron2018invariant}.

For evaluating explanation accuracy, \emph{Lamole} is compared with three types of alternative methods: 1) GCN with feature-based explainability techniques, including SmoothGrad~\cite{smoothgrad}, GradInput~\cite{gradinput}, and GradCAM~\cite{Selvaraju_2017_ICCV}, 2) Bert with the above feature-based explainability techniques, where Group SELFIES is used as input for a fair comparison, and 3) explainable GNNs including OrphicX~\cite{lin2022orphicx}, GNNExplainer~\cite{gnnexplainer}, PGExplainer~\cite{pgexplainer}, and Gem~\cite{lin2021generative}. The details of parameter settings can be found in the supplemental material.


\setlength{\tabcolsep}{8pt}
\begin{table*}[t]
  \centering
  \caption{Mean Classification Accuracy on the Seven Datasets (\%)}
  \begin{threeparttable}  
    \begin{tabular}{cccccccc}
    \toprule
    Methods & Mutag & Mutagen & PTC-FM & PTC-FR & PTC-MM & PTC-MR & Liver \\
    \midrule
    GCN~\cite{kipf2016semi} & 84.6  & 78.9  & 54.8  & 63.0  & 57.8  & 53.3  & 41.1 \\
    DGCNN~\cite{zhang2018end} & 85.8  & 74.8  & 57.3  & 63.5  & 61.0  & 58.6  & 44.6 \\
    edGNN~\cite{jaume2019edgnn} & 86.9  & 75.2  & 59.8  & 65.7  & \boldmath{}\textbf{64.4$^{\dagger}$}\unboldmath{} & 56.3  & 44.5 \\
    GIN~\cite{xu2018powerful} & 87.5  & \boldmath{}\textbf{82.3$^{\ddagger}$}\unboldmath{} & \boldmath{}\textbf{62.1$^{\dagger}$}\unboldmath{} & 66.2  & \boldmath{}\textbf{65.1$^{\ddagger}$}\unboldmath{} & 64.0  & 44.9 \\
    RW-GNN~\cite{nikolentzos2020random} & 87.2  & 80.3  & 61.9  & 64.0  & 62.4  & 57.0  & 43.2 \\
    DropGNN~\cite{papp2021dropgnn} & \boldmath{}\textbf{89.4$^{\ddagger}$}\unboldmath{} & \boldmath{}\textbf{80.7$^{\dagger}$}\unboldmath{} & 62.0  & 66.0  & 63.7  & 64.2  & 45.0 \\
    IEGN~\cite{maron2018invariant} & 84.6  & 80.1  & 60.8  & 59.8  & 61.1  & 59.5  & 45.3 \\
    ChemBERTa~\cite{ChemBERTa} & 86.8 &  78.0 & 60.0  & 65.7  & 60.4  & 58.7  & 45.7 \\
    \midrule
    \emph{Lamole}+DistilBert & 84.2  & 76.8  & 57.5  & 69.0  & 60.2  & \boldmath{}\textbf{64.5$^{\dagger}$}\unboldmath{} & \boldmath{}\textbf{47.2$^{\dagger}$}\unboldmath{} \\
    \emph{Lamole}+DeBerta & 86.8  & 73.7  & 58.6  & \boldmath{}\textbf{69.5$^{\dagger}$}\unboldmath{} & 59.7  & 63.8  & 45.8 \\
    \emph{Lamole}+Bert & \boldmath{}\textbf{88.2$^{\dagger}$}\unboldmath{} & 74.5  & \boldmath{}\textbf{62.4$^{\ddagger}$}\unboldmath{} & \boldmath{}\textbf{70.0$^{\ddagger}$}\unboldmath{} & 61.2  & \boldmath{}\textbf{66.0$^{\ddagger}$}\unboldmath{} & \boldmath{}\textbf{47.5$^{\ddagger}$}\unboldmath{} \\
    \bottomrule
    \end{tabular}%
     \begin{tablenotes}
        \footnotesize
        \item $^{\ddagger}$ and $^{\dagger}$ denote the best and the second-best results, respectively.
      \end{tablenotes}
  \end{threeparttable}
  \label{tab:classfication}%
\end{table*}%

\begin{table*}[t]
  \centering
  \caption{Mean Explanation Accuracy on the Seven Datasets (\%)}
    \begin{tabular}{cccccccc}
        \toprule
    Methods & Mutag & Mutagen & PTC-FM & PTC-FR & PTC-MM & PTC-MR & Liver \\
    \midrule
    GradInput+GCN~\cite{gradinput} & 70.3  & 67.9  & 69.7  & 66.4  & 64.6  & 65.0  & 73.0 \\
    GradCAM+GCN~\cite{Selvaraju_2017_ICCV} & 69.8  & 67.0  & 71.0  & 67.9  & 66.2  & 67.3  & 69.4 \\
    SmoothGrad+GCN~\cite{smoothgrad} & 69.2  & 66.8  & 67.5  & 62.6  & 64.9  & 63.1  & 66.4 \\
    \midrule
    GradInput+Bert~\cite{gradinput} & 75.1  & 72.6  & 73.0  & 68.9  & 65.6  & 69.6  & 75.6 \\
    GradCAM+Bert~\cite{Selvaraju_2017_ICCV} & 75.3  & 72.4  & 77.5  & 70.0  & 70.2  & 73.0  & 76.0 \\
    SmoothGrad+Bert~\cite{smoothgrad} & 73.4  & 72.8  & 73.7  & 71.0  & 67.0  & 69.9  & 75.1 \\
    \midrule
    GNNExplainer~\cite{gnnexplainer} & 70.6  & 64.2  & 68.9  & 67.9  & 66.8  & 67.1  & 72.1 \\
    PGExplainer~\cite{pgexplainer} & 66.5  & 58.7  & 70.3  & 68.0  & 65.9  & 67.0  & 71.5 \\
    Gem~\cite{lin2021generative} & 73.7  & 66.0  & 71.3  & 69.0  & 68.9  & 69.2  & 73.6 \\
    OrphicX~\cite{lin2022orphicx} & \boldmath{}\textbf{78.0$^{\ddagger}$}\unboldmath{} & 71.5  & 74.6  & 70.4  & \boldmath{}\textbf{70.9$^{\dagger}$}\unboldmath{} & 71.4  & 74.0 \\
    \midrule
    \emph{Lamole}+DistilBert & 70.9  & 73.0  & 74.0  & 70.2  & 69.6  & \boldmath{}\textbf{78.1$^{\ddagger}$}\unboldmath{} & \boldmath{}\textbf{76.1$^{\ddagger}$}\unboldmath{} \\
    \emph{Lamole}+DeBerta & 76.1  & \boldmath{}\textbf{75.0$^{\dagger}$}\unboldmath{} & \boldmath{}\textbf{79.9$^{\dagger}$}\unboldmath{} & \boldmath{}\textbf{72.1$^{\dagger}$}\unboldmath{} & 70.3  & \boldmath{}\textbf{77.2$^{\dagger}$}\unboldmath{} & 75.0 \\
    \emph{Lamole}+Bert & \boldmath{}\textbf{77.8$^{\dagger}$}\unboldmath{} & \boldmath{}\textbf{75.2$^{\ddagger}$}\unboldmath{} & \boldmath{}\textbf{81.1$^{\ddagger}$}\unboldmath{} & \boldmath{}\textbf{72.2$^{\ddagger}$}\unboldmath{} & \boldmath{}\textbf{72.0$^{\ddagger}$}\unboldmath{} & 73.1  & \boldmath{}\textbf{77.3$^{\dagger}$}\unboldmath{} \\
    \bottomrule
    \end{tabular}%
      \begin{tablenotes}
        \footnotesize
        \item $^{\ddagger}$ and $^{\dagger}$ denote the best and the second-best results, respectively.
      \end{tablenotes}
  \label{tab:explanationAUC}%
\end{table*}%

\begin{table*}[htbp]
  \centering
  \caption{MEAN EXPLANATIONS' FIDELITY ON THE SEVEN DATASETS}
    \begin{tabular}{cccccccc}
    \toprule
    Methods & Mutag & Mutagen & PTC-FM & PTC-FR & PTC-MM & PTC-MR & Liver \\
    \midrule
    SmoothGrad+GCN~\cite{smoothgrad} & 0.31 & 0.28 & 0.30 & 0.25 & 0.29 & 0.27 & 0.33 \\
    CAM+GCN~\cite{cam} & 0.30 & 0.27 & 0.29 & 0.26 & 0.28 & 0.32 & 0.35 \\
    GradInput+GCN~\cite{gradinput} & 0.32 & 0.26 & 0.31 & 0.28 & 0.25 & 0.26 & 0.34 \\
    GradCAM+GCN~\cite{Selvaraju_2017_ICCV} & 0.33 & 0.29 & 0.34 & 0.30 & 0.31 & 0.33 & 0.36 \\
    \midrule
    SmoothGrad+Bert~\cite{smoothgrad} & 0.40 & 0.38 & 0.42 & 0.39 & 0.35 & 0.38 & 0.45 \\
    CAM+Bert~\cite{cam} & 0.39 & 0.37 & 0.36 & 0.38 & 0.34 & 0.37 & 0.43 \\
    GradInput+Bert~\cite{gradinput} & 0.42 & 0.40 & 0.41 & 0.36 & 0.35 & 0.39 & 0.46 \\
    GradCAM+Bert~\cite{Selvaraju_2017_ICCV} & 0.43 & 0.41 & 0.47 & 0.40 & 0.42 & 0.44 & 0.48 \\
    OrphicX~\cite{lin2022orphicx} & 0.45 & 0.43 & 0.44 & 0.40 & 0.41 & 0.43 & 0.47 \\
    GNNExplainer~\cite{gnnexplainer} & 0.35 & 0.31 & 0.34 & 0.32 & 0.33 & 0.35 & 0.40 \\
    PGExplainer~\cite{pgexplainer} & 0.33 & 0.29 & 0.35 & 0.31 & 0.32 & 0.34 & 0.39 \\
    Gem~\cite{lin2021generative} & 0.41 & 0.36 & 0.38 & 0.37 & 0.39 & 0.40 & 0.44 \\
    \midrule
    \emph{Lamole}+DistilBert & 0.50 & 0.52 & 0.55 & 0.51 & 0.53 & \textbf{0.62}$^{\ddagger}$ & 0.58 \\
    \emph{Lamole}+DeBerta & \textbf{0.58}$^{\dagger}$ & \textbf{0.60}$^{\dagger}$ & \textbf{0.65}$^{\dagger}$ & \textbf{0.55}$^{\dagger}$ & \textbf{0.56}$^{\dagger}$ & \textbf{0.61}$^{\dagger}$ & \textbf{0.59}$^{\dagger}$ \\
    \emph{Lamole}+Bert & \textbf{0.61$^{\ddagger}$} & \textbf{0.62}$^{\ddagger}$ & \textbf{0.68}$^{\ddagger}$ & \textbf{0.57}$^{\ddagger}$ & \textbf{0.58}$^{\ddagger}$ & 0.57 & \textbf{0.63}$^{\ddagger}$ \\
    \bottomrule
    \end{tabular}%
  \label{tab:fidelity_comparison}%
\end{table*}%

\subsection{Results and Discussion}
\textbf{Prediction Performance}. Table \ref{tab:classfication} shows the classification accuracy of the compared algorithms. As we can see, our proposed \emph{Lamole}+DistilBert, \emph{Lamole}+DeBerta, and \emph{Lamole}+Bert not only can provide explainability but also can achieve comparable prediction accuracy as compared to existing predictive methods. In addition, \emph{Lamole} models show superior performance over ChemBERTa. This suggests that using molecular representations with more chemical semantics, like Group SELFIES, can help LMs better learn the chemical semantics and structure-property relationships.

\textbf{Explanation Performance.} Table \ref{tab:explanationAUC} presents the explanation accuracy of the compared explainability techniques. It should be noted that the ground truth annotations used in our work provide additional supervisory signals. Therefore, we also align the generated explanations of these baselines with the annotations for fair comparison, and the ground truth annotation rate is 10\%.
\emph{Lamole} improves the explanation accuracy by $1.4\%\sim 14.3\%$ compared to the baseline methods. 

The results of the explanation accuracy of \textit{Lamole} under different annotation rates (10\%, 20\%, 50\%, and 100\%) on the PTC-FR and PTC-MM datasets are shown in Fig. \ref{ratemutag} and Fig. \ref{ratefr} (a), (b). It is clear that more annotations can constantly enhance the accuracy of the explanation. Compared to \emph{Lamole} ($-\mathcal{L}_M$), only using 10\% molecules with ground truth annotations can significantly improve explanation accuracy by up to 5\%. Using more annotations (from 10\% to 20\%) can achieve significant improvement in explanation accuracy. However, raising the rate from 50\% to 100\% can bring a limited increase in explanation accuracy on the four PTC datasets. Therefore, there is a trade-off between the explanation accuracy and additional annotation costs.

The fidelity results across the seven datasets presented in Table \ref{tab:fidelity_comparison} confirm that our method, {\em Lamole}, generates explanations that are more faithful to the underlying model prediction than all baselines. The best performing {\em Lamole}+Bert, consistently achieved the highest scores on nearly all benchmarks, including 0.61 on Mutag, 0.62 on Mutagen, and 0.68 on PTC-FM. This performance is substantially higher compared to dedicated explainers like $\text{OrphicX}$ and gradient methods. The data indicate that {\em Lamole}'s integration of attention and gradient signals, combined with its functional-group focus, produces explanations where the removal of the identified fragments causes a maximal reduction in the model's prediction score. This confirms the reliability of our approach.

\begin{figure*}[t]
	\centering  
	\subfigure[Mutag]{
		\includegraphics[width=0.32\linewidth]{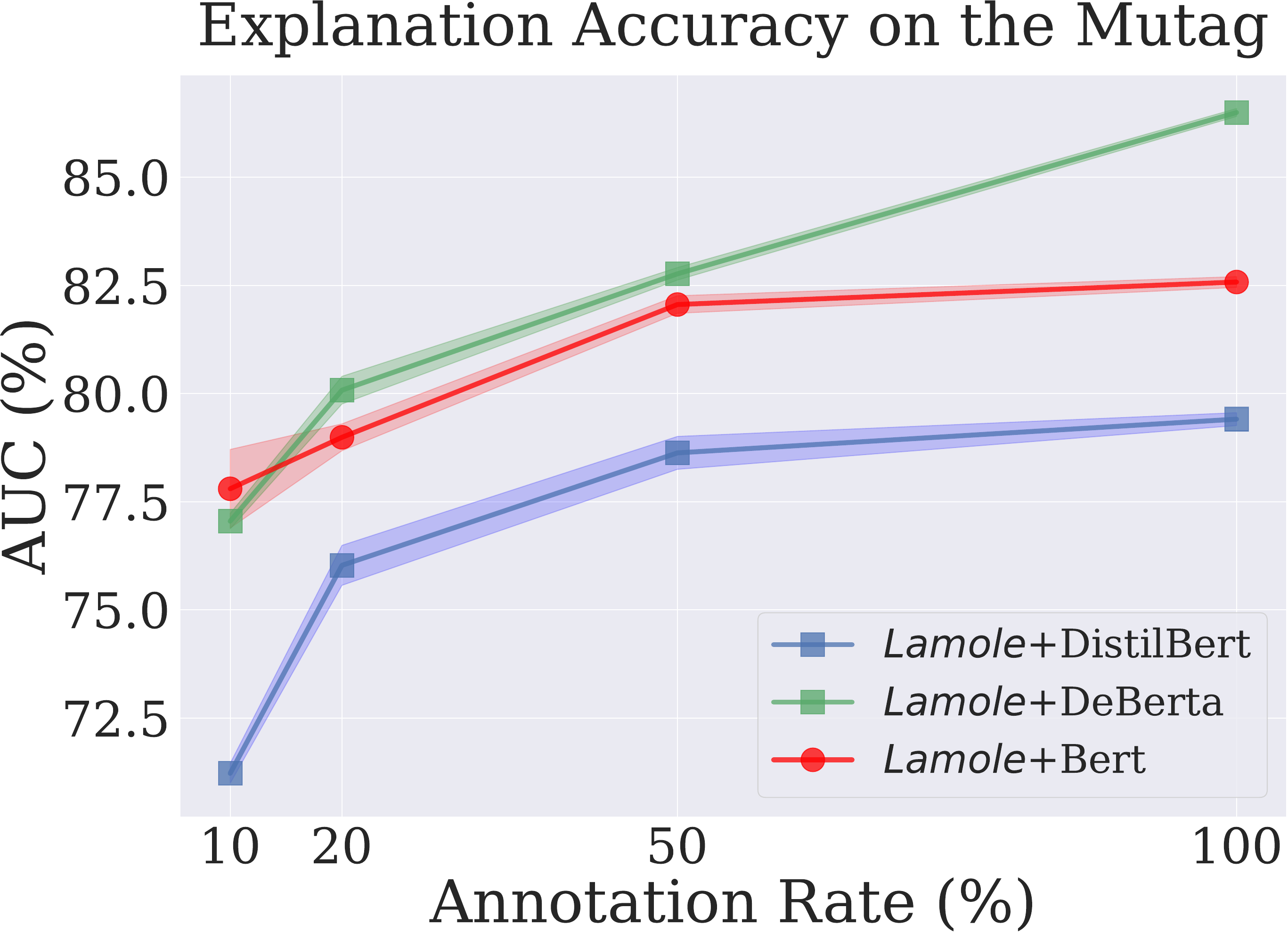}}
	\subfigure[PTC-FM]{
		\includegraphics[width=0.32\linewidth]{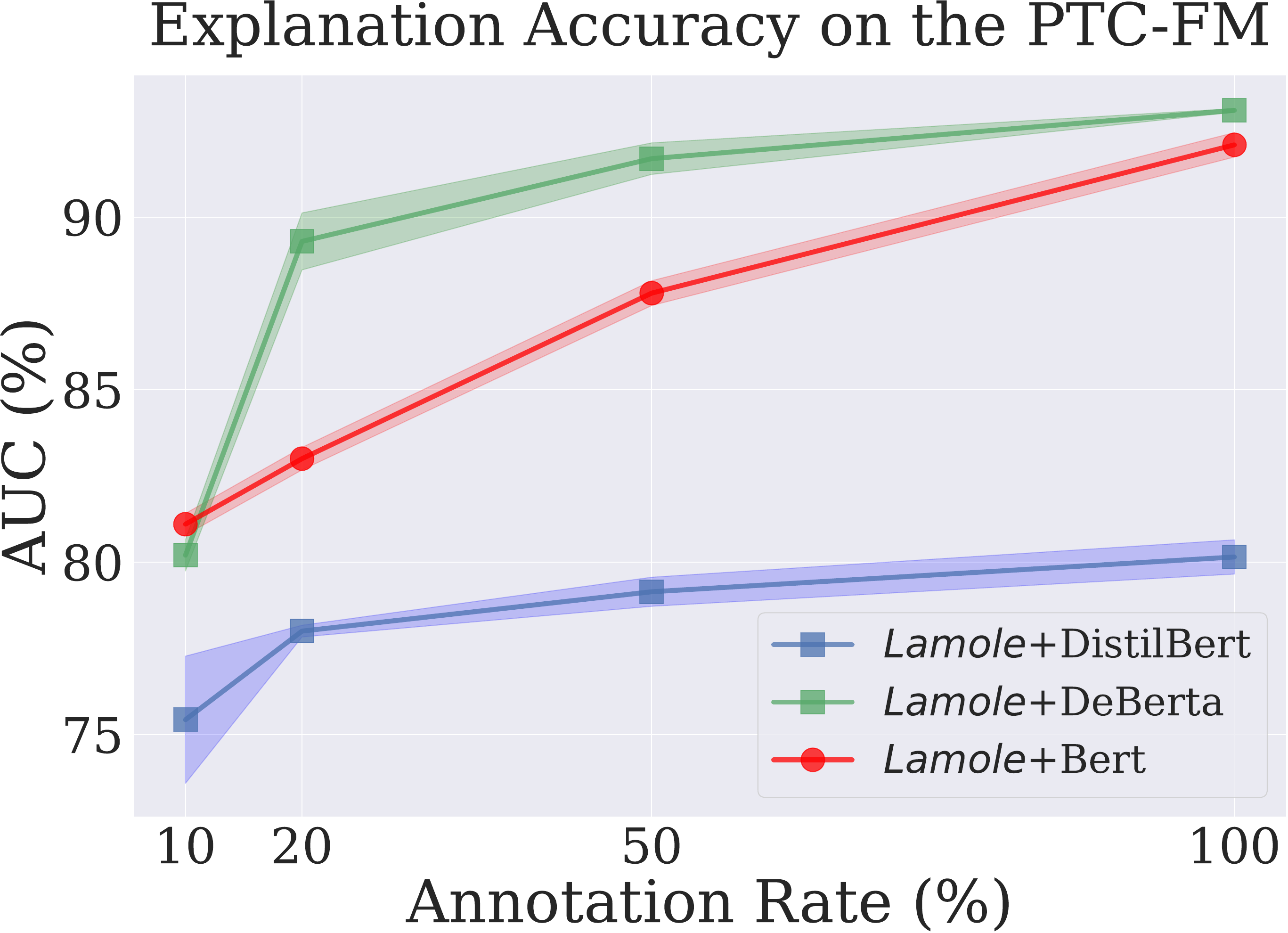}}
	\subfigure[PTC-MR]{
		\includegraphics[width=0.32\linewidth]{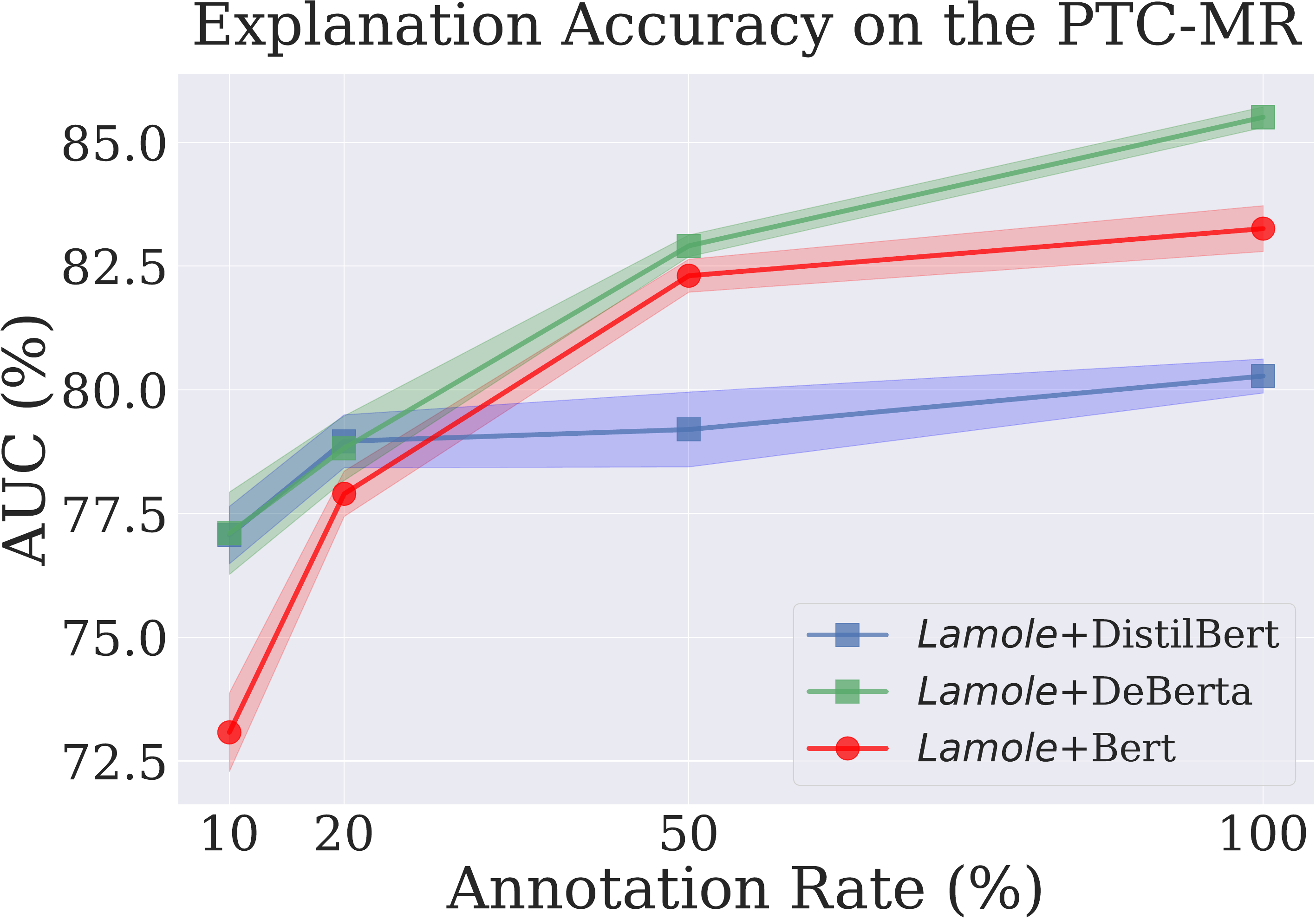}}
	\caption{The explanation accuracy of \textit{Lamole} with different annotation rates on the Mutag, PTC-FM, and PTC-MR datasets.}
  \label{ratemutag}
\end{figure*}

\begin{figure*}[t]
	\centering  
	\subfigure[PTC-FR]{
		\includegraphics[width=0.255\linewidth]{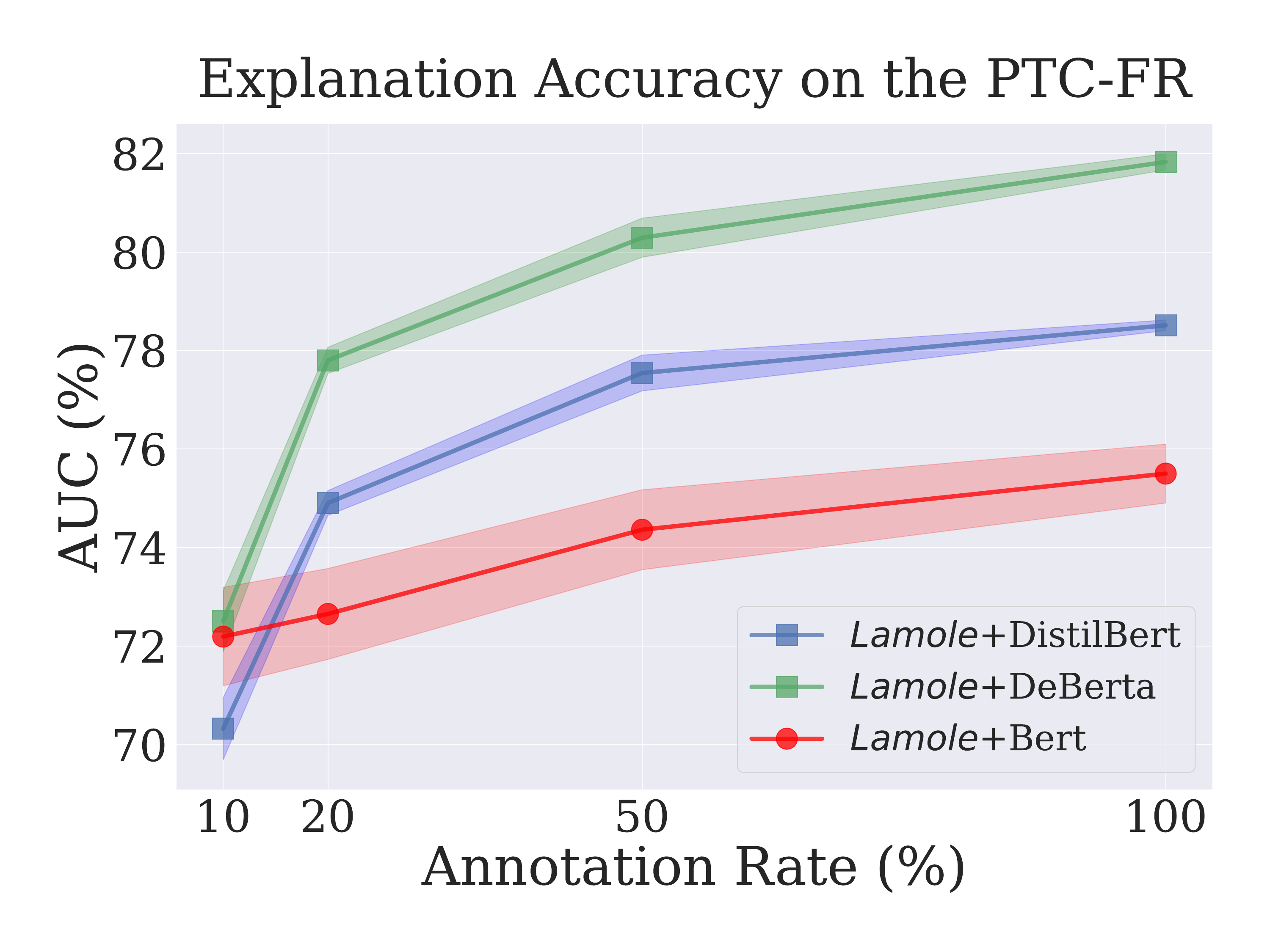}}
	\subfigure[PTC-MM]{
		\includegraphics[width=0.255\linewidth]{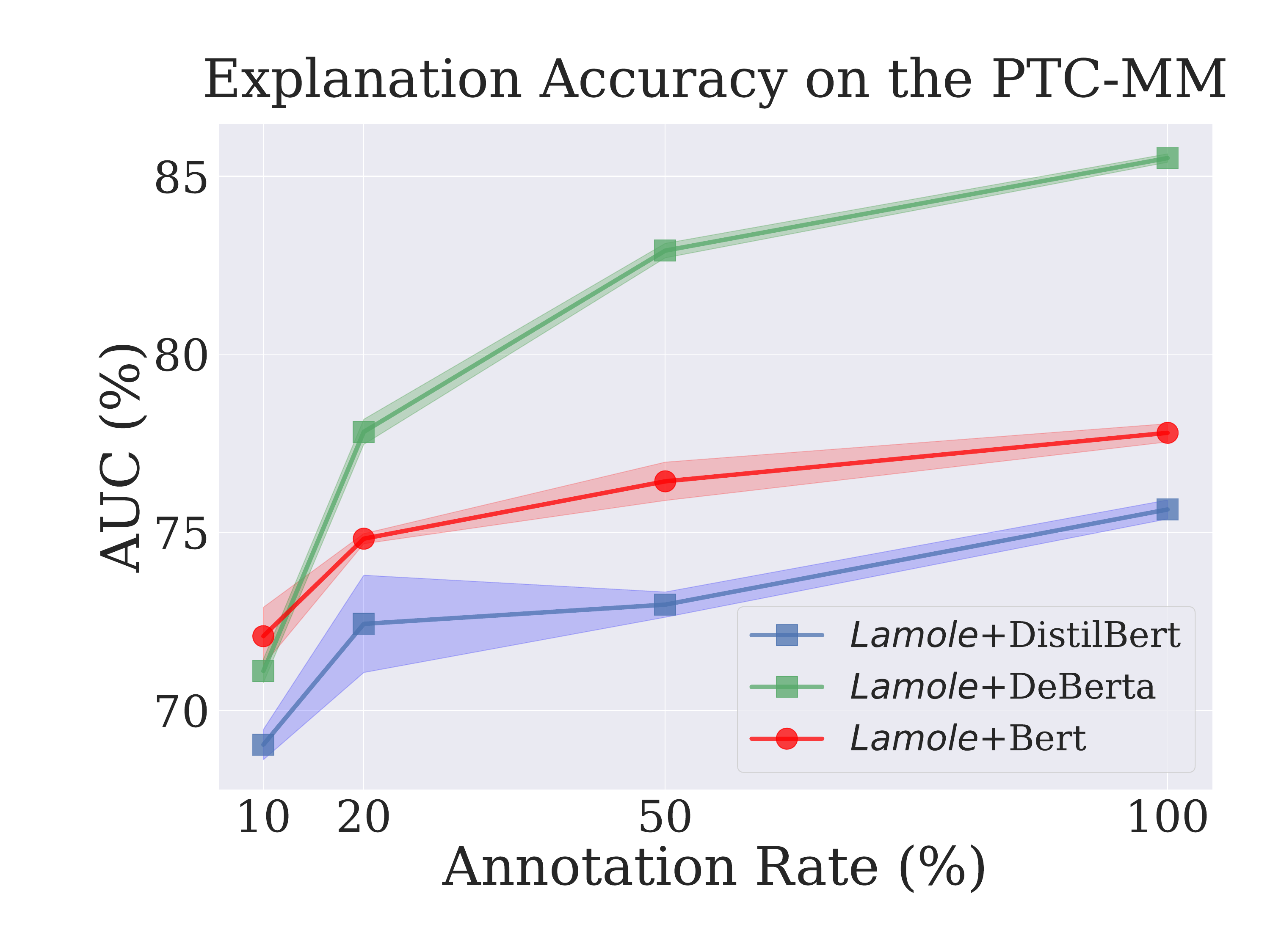}}
        \subfigure[Molecule (ID: 155)]{
            \includegraphics[width=0.45\linewidth]{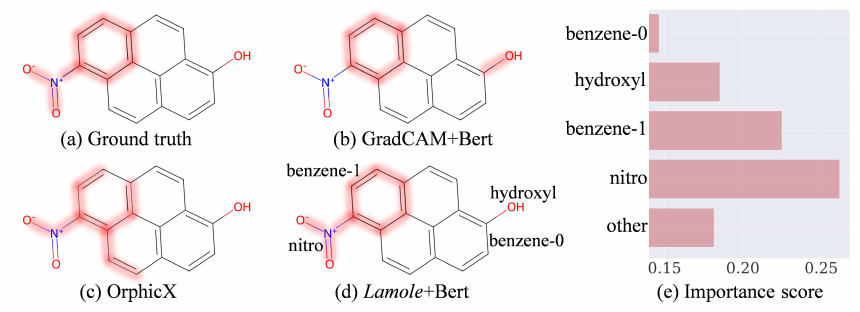}
        }
        
	\caption{The explanation accuracy of \textit{Lamole} with different annotation rates on the PTC-FR and PTC-MM datasets, and the explanation visualization of one molecule from the Mutag dataset.}
  \label{ratefr}
\end{figure*}



In addition, we discuss the rationale of using labeled annotations. 
\emph{Lamole} requires human-labeled annotations. Due to the huge knowledge base of LLMs, we explored the use of LLMs, including ChatGPT and ChemLLM~\cite{zhang2024chemllm}, to annotate the ground truth. We input molecules' SMILES strings into the two LLMs to ask the ground truth. The explanation accuracy results are shown in Table \ref{tab:llmlable}.

\setlength{\tabcolsep}{4pt}
\begin{table}[t]\scriptsize
  \centering
  \caption{The explanation accuracy results when using different annotation methods}
    \begin{tabular}{ccccccc}
    \toprule
    Methods & Mutag & PTC-FM & PTC-FR & PTC-MM & PTC-MR & Liver \\
    \midrule
    \textit{\emph{Lamole} (ChatGPT)} & \textcolor[rgb]{ .2,  .2,  .2}{67.6} & 64.5  & 55.0  & 51.0  & 65.6  & 72.7 \\
    \textit{\emph{Lamole} (ChemLLM)} & 62.5  & 72.1  & 61.5  & 57.1  & 71.9  & 72.7 \\
    \textit{\emph{Lamole} (Human)} & \textbf{77.8} & \textbf{81.1} & \textbf{72.2} & \textbf{72.0} & \textbf{73.1} & \textbf{77.3} \\
    \bottomrule
    \end{tabular}%
  \label{tab:llmlable}%
\end{table}%

It is obvious that there is a significant decrease in explanation accuracy, indicating that existing LLMs may make incorrect annotations. Therefore, we argue that this human-in-the-loop strategy — providing slight human annotations — to guide learning is reasonable and necessary for critical scientific domains.


We selected some representative molecules for explanation visualization. These explanations are shown in Figs. \ref{fig:intro}, \ref{ratefr} (c), \ref{fig:vis2}, and Fig. S-1 (see supplemental material), respectively. The right panel of those figures is the importance scores obtained by \emph{Lamole}, where "other" in the figures is the average importance score of other unlisted functional groups/fragments. 
Compared to baseline methods, \emph{Lamole} provides chemically meaningful explanations. Particularly, the interaction among the functional groups is successfully captured. 
As shown in Figs. \ref{fig:vis2} (b) and S-1 (a), \emph{Lamole} accurately and confidently identify benzene with amido group and benzene-1 with nitro-1 group as explanations, respectively. While other methods can neither provide chemically meaningful explanations nor reflect the functional group interactions. These explanations demonstrate \emph{Lamole}’s superior interpretation in faithfully revealing the structure-property relationships. However, as depicted in Fig. S-1 (b), although the ground truth substructures, i.e., benzene-1 and amido ground, are identified, other functional groups/fragments such as carbonyl-0, $\mathrm{Br}$-0, and $\mathrm{Br}$-1 also have relatively higher importance scores. This may be due to complex interactions caused by multiple functional groups. In future work, more strategies may need to be designed to reveal such complex functional group interactions.

\begin{figure*}[t]
    \centering
    \subfigure[Molecule (ID: 156) from the Mutag dataset]{
        \includegraphics[width=0.48\linewidth]{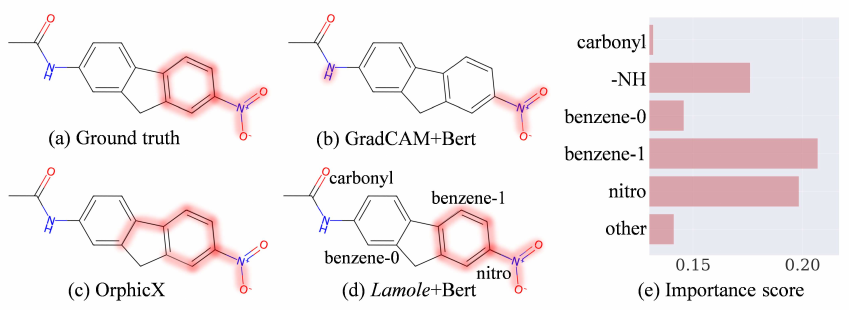}
    }
    \subfigure[Molecule (ID: 574) from the Liver dataset]{
        \includegraphics[width=0.48\linewidth]{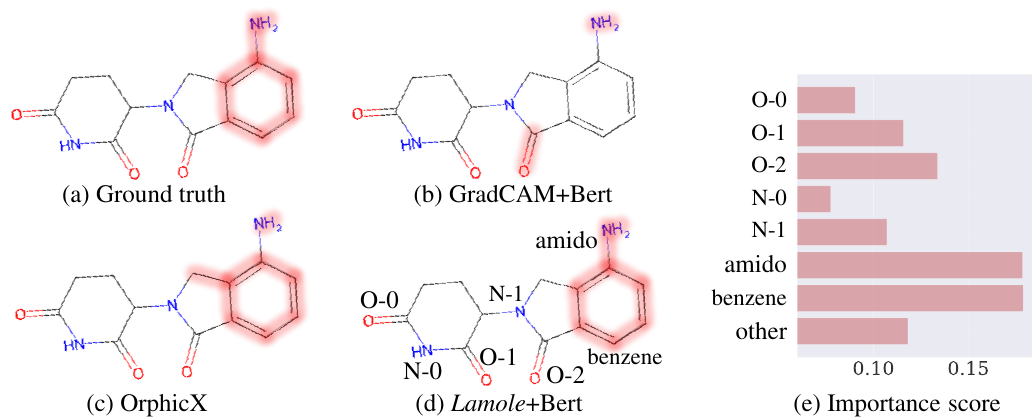}
    }
    \caption{Explanation visualization molecules from the Mutag and the Liver datasets.}

    \label{fig:vis2}
\end{figure*}

In addition, we evaluated the performance of the compared algorithms by using the proposed explanation plausibility metric $\mathrm{EP}$. The statistical results of $\mathrm{EP}$ on the PTC-FR, PTC-MM, and Liver datasets are presented in Figs. \ref{epmutag} and S-2.
From the figures, we can observe that the $\mathrm{EP}$ values of the comparison algorithm are slightly lower, which means that the algorithms cannot confidently reflect the relationships between structure and property.
Compared to the comparison algorithm, the $\mathrm{EP}$ values of \emph{Lamole} have increased by $2\%\sim9\%$. 
The explanations' plausibility $\mathrm{EP}(g)$ is defined as the ratio of the difference between the mean importance scores of ground truth and the mean importance scores of non-ground truth to the mean importance scores of non-ground truth. A larger ratio indicates that ground truth's importance scores exceed non-ground truth's. The high $\mathrm{EP}(g)$ values of \textit{Lamole} indicate \emph{Lamole} can improve the importance scores of ground truth and suppress the importance scores of non-ground truth, leading to higher confidence in matching the ground truth.

\begin{figure*}[t]
	\centering  
	\subfigure[Mutag]{
		\includegraphics[width=0.32\linewidth]{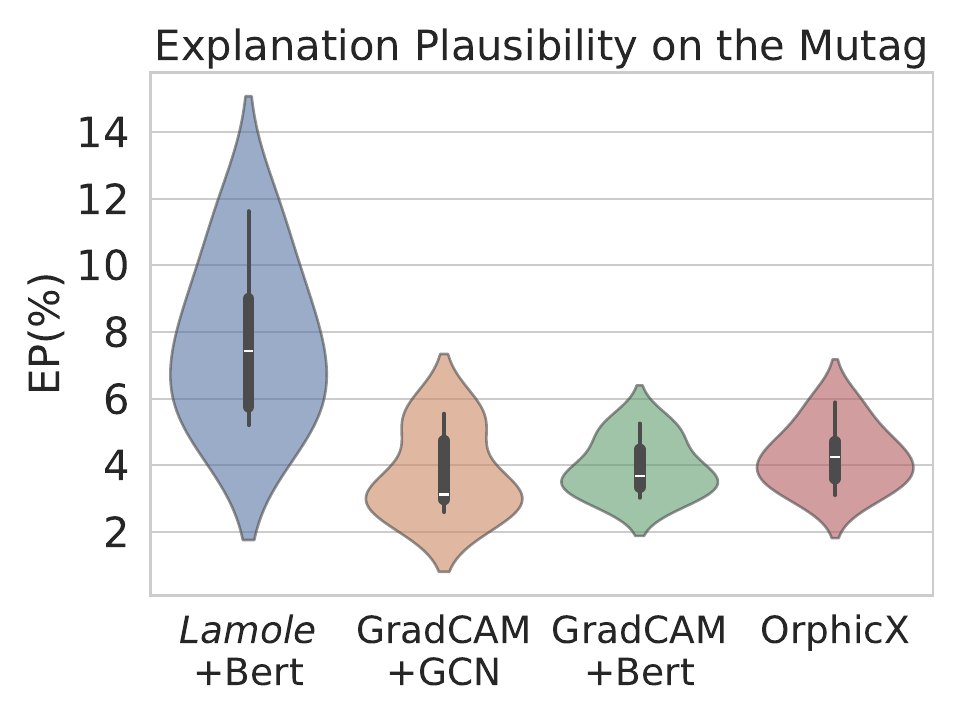}}
	\subfigure[PTC-FM]{
		\includegraphics[width=0.32\linewidth]{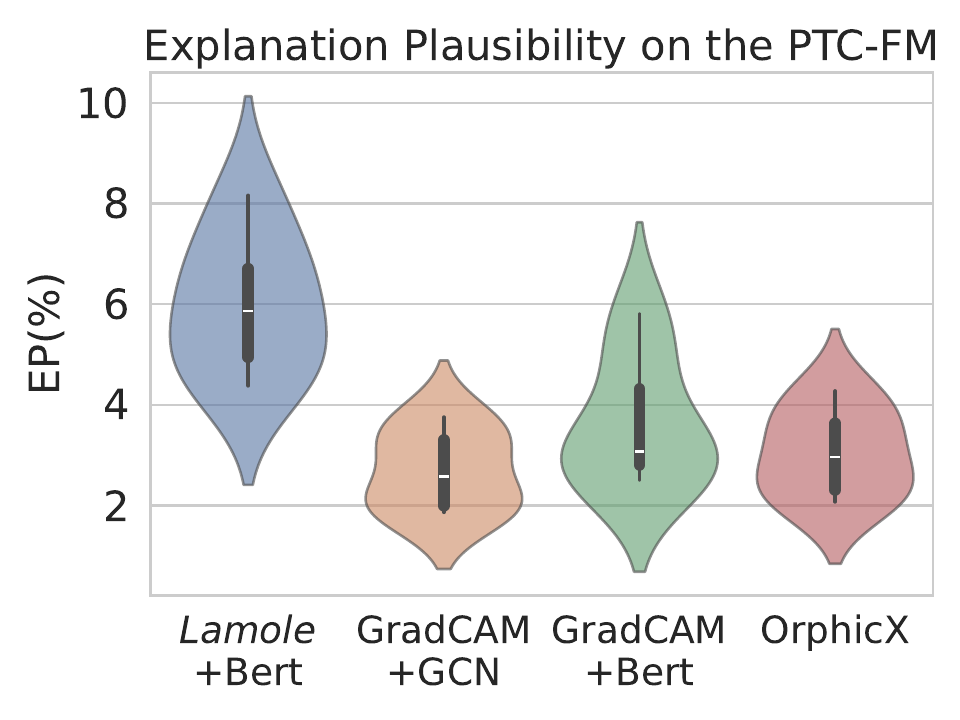}}
	\subfigure[PTC-MR]{
		\includegraphics[width=0.32\linewidth]{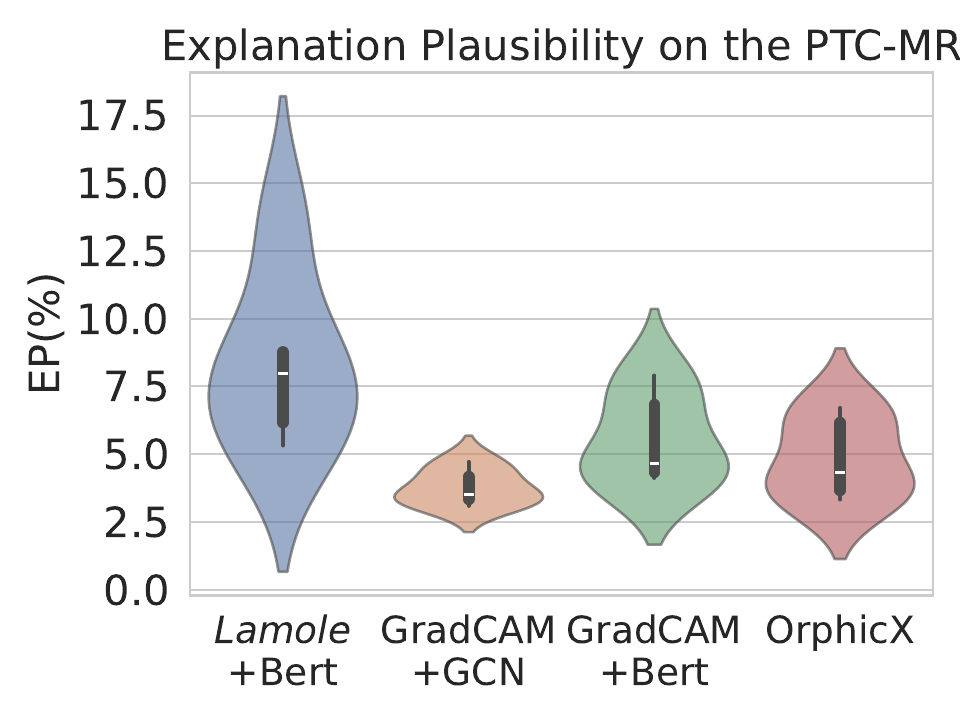}}
	\caption{The explanation plausibility of the compared algorithms on the Mutag, PTC-FM, and PTC-MR datasets.}
  \label{epmutag}
\end{figure*}

\subsection{Applicability to Broader Properties}

The above experiments focus on toxicity-related properties where some ground-truth concepts are well-understood. To further investigate whether the explainable prediction ability applies to broader properties, we have performed additional experiments on two critical ADME properties: solubility and permeability.

Since large-scale human-annotated datasets for ADME properties are unavailable, we employ an automated, descriptor-driven approach to generate the ground-truth functional group labels required for evaluating explanation accuracy on the ZINC dataset. We leverage established physicochemical principles quantified by the $\text{RDKit}$ cheminformatics package~\cite{landrum2013rdkit} to define chemically meaningful key fragments. For both solubility and permeability, the process hinges on quantifying local contributions to defining global descriptors, such as topological Polar Surface Area ($\text{tPSA}$) for polarity and LogP for hydrophobicity. We utilize $\text{RDKit}$ to calculate the contribution of each atom to the relevant property descriptor. We determine the ground truth structures by aggregating these atomic contributions: the contributions of all atoms belonging to a specific Group SELFIES fragment are summed. The fragments whose summed contributions are among the highest are then designated as the ground truth structures. Fragments with the highest $\text{tPSA}$ are labeled as ground truth structures \footnote{\textcolor{black}{It should be noted that the purpose of using this RDKit-based annotation is to provide computational references, rather than the true chemical causality --- this needs to be rigorously verified through wet experiments.}}.

The results presented in Table \ref{tab:predictive_rmse} and Table \ref{tab:explanation_comparison} demonstrate that the advantages of the {\em Lamole} framework extend to these more diverse and complex pharmacological tasks, confirming its applicability beyond toxicity prediction. Regarding predictive performance (Table \ref{tab:predictive_rmse}), {\em Lamole} shows substantial superiority over GNN baselines in terms of Root Mean Square Error ($\text{RMSE}$). For instance, {\em Lamole}+Bert achieved an $\text{RMSE}$ of 0.65 for solubility. A similar trend holds for permeability, where {\em Lamole}+Bert achieved an $\text{RMSE}$ of 2.1, significantly outperforming GNN baselines. This predictive accuracy confirms that {\em Lamole}'s architecture captures the subtle structure-activity relationships crucial for accurate $\text{ADME}$ prediction.

The results for explanation quality (Table \ref{tab:explanation_comparison}) further show the method's generalizability by yielding higher explanation accuracy and fidelity scores. The framework achieved the highest explanation accuracy ($\text{Exp. Acc.}$) for both solubility (89.1\%) and permeability (73.5\%), indicating that {\em Lamole} is superior at identifying the core functional groups that govern physicochemical behavior. Furthermore, the fidelity scores—which measure how faithful the explanation is to the model's prediction—are also highest for {\em Lamole} variants. The explanation visualization shown in Fig. S-3 and the explanation plausibility shown in Fig. S-4 also confirmed the accuracy of the derived explanations.

\begin{table}[htbp]\scriptsize
  \centering
  \caption{Comparison of Prediction Performance (RMSE) on Solubility and Permeability}
    \begin{tabular}{ccc}
    \toprule
    Methods & Solubility & Permeability \\
    \midrule
    GCN~\cite{kipf2016semi} & 0.88 & 3.5 \\
    DGCNN~\cite{zhang2018end} & 0.85 & 3.4 \\
    edGNN~\cite{jaume2019edgnn} & 0.82 & 3.3 \\
    GIN~\cite{xu2018powerful} & 0.79 & 3.1 \\
    RW-GNN~\cite{nikolentzos2020random} & 0.77 & 3.0 \\
    DropGNN~\cite{papp2021dropgnn} & 0.75 & 2.9 \\
    IEGN~\cite{maron2018invariant} & 0.73 & 2.9 \\
    \midrule
    \emph{Lamole}+DistilBert & \textbf{0.70}$^{\dagger}$ & 2.8 \\
    \emph{Lamole}+DeBerta & 0.74 & \textbf{2.5}$^{\dagger}$ \\
    \emph{Lamole}+Bert & \textbf{0.65}$^{\ddagger}$ & \textbf{2.1}$^{\ddagger}$ \\
    \bottomrule
    \end{tabular}%
  \label{tab:predictive_rmse}%
\end{table}%

\begin{table}[htbp]
  \centering
  \caption{Comparison of Explanation Quality on Solubility and Permeability}
   \resizebox{0.99\linewidth}{!}{
    \begin{tabular}{ccccc}
    \toprule
    \multirow{2}[4]{*}{Methods} & \multicolumn{2}{c}{Solubility} & \multicolumn{2}{c}{Permeability} \\
\cmidrule{2-5}    & Exp. Acc. ($\%$) & Fidelity & Exp. Acc. ($\%$) & Fidelity \\
    \midrule
    SmoothGrad+GCN~\cite{smoothgrad} & 78.5 & 0.35 & 64.1 & 12.3 \\
    CAM+GCN~\cite{cam} & 79.1 & 0.38 & 65.0 & 13.5 \\
    GradInput+GCN~\cite{gradinput} & 77.8 & 0.32 & 63.8 & 11.9 \\
    GradCAM+GCN~\cite{Selvaraju_2017_ICCV} & 80.3 & 0.41 & 66.2 & 14.0 \\
    SmoothGrad+Bert~\cite{smoothgrad} & 81.5 & 0.43 & 67.5 & 14.5 \\
    CAM+Bert~\cite{cam} & 82.2 & 0.45 & 68.1 & 14.8 \\
    GradInput+Bert~\cite{gradinput} & 81.0 & 0.42 & 66.9 & 14.1 \\
    GradCAM+Bert~\cite{Selvaraju_2017_ICCV} & 83.0 & 0.46 & 69.5 & 15.1 \\
    \midrule
    OrphicX~\cite{lin2022orphicx} & 84.5 & 0.48 & 70.0 & 15.5 \\
    GNNExplainer~\cite{gnnexplainer} & 85.1 & 0.49 & 70.8 & 15.8 \\
    PGExplainer~\cite{pgexplainer} & 83.9 & 0.47 & 69.8 & 15.3 \\
    Gem~\cite{lin2021generative} & 85.5 & 0.49 & 71.0 & 16.0 \\
    \midrule
    \emph{Lamole}+DistilBert & 88.4 & \textbf{0.52}$^{\dagger}$ & 72.1 & \textbf{16.5}$^{\dagger}$ \\
    \emph{Lamole}+DeBerta & \textbf{88.7}$^{\dagger}$ & 0.50 & \textbf{73.2}$^{\dagger}$ & 15.9 \\
    \emph{Lamole}+Bert & \textbf{89.1}$^{\ddagger}$ & \textbf{0.56}$^{\ddagger}$ & \textbf{73.5}$^{\ddagger}$ & \textbf{17.6}$^{\ddagger}$ \\
    \bottomrule
    \end{tabular}%
    }
  \label{tab:explanation_comparison}%
\end{table}%




\subsection{Ablation Studies}

We conducted ablation studies for each component in our \emph{Lamole}. Specifically, we removed the attention weights in the explanations, removed the marginal loss, and only used attention weights as explanations. The corresponding ablation algorithms are named \emph{Lamole} ($-\alpha$), \emph{Lamole} ($-\mathcal{L}_M$), and \emph{Lamole} (Att), respectively. The results are shown in Fig. \ref{fig:nipsablation}. The explanation accuracy decreases by 1.4\%$\sim$2.3\% when removing the attention weights. Removing the marginal loss can decrease the explanation accuracy by 1.0\%$\sim$5.0\%. Regarding the results of using only attention weights, the explanation accuracy decreases by 1.4\%$\sim$6.3\%. The above results confirm the effectiveness of using the marginal loss, attention weights, and gradients.



\begin{figure*}[t] 
  \centering   
  \includegraphics[width=0.88\linewidth]{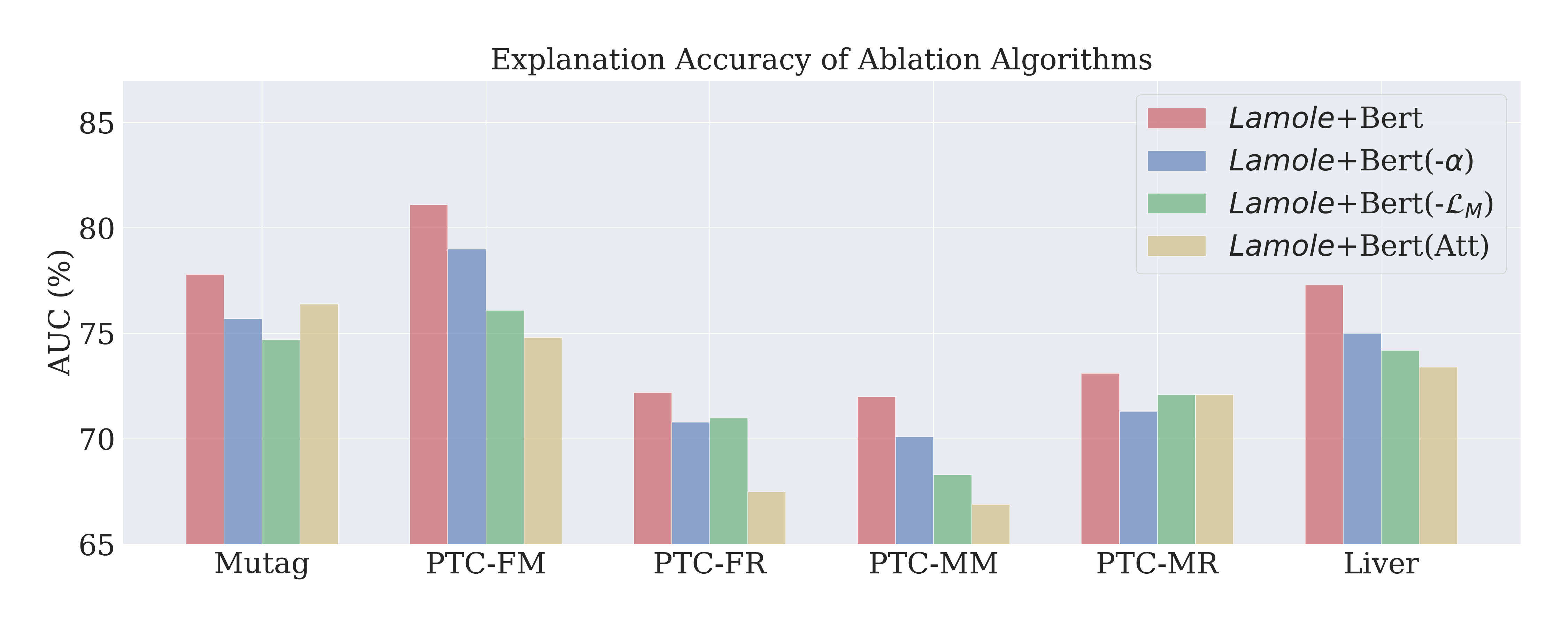}
    \vspace{-0.5cm}
  \caption{The explanation accuracy of $\textit{Lamole}$+Bert, $\textit{Lamole}$+Bert ($-\alpha$), $\textit{Lamole}$+Bert ($-\mathcal{L}_M$), and $\textit{Lamole}$ (Att).}
  \label{fig:nipsablation}
\end{figure*}

From the perspective of model training, the marginal loss enables the model to be trained under the causal signals. To verify, we compared the classification performance with and without the marginal loss, as shown in Table \ref{tab:woLMclass}. Without the marginal loss, the classification accuracy degrades by 0.7\%$\sim$3.7\%. The above results indicate that marginal loss could help identify the causal features, thereby improving classification accuracy.

\setlength{\tabcolsep}{3pt}
\begin{table}[t]\scriptsize
  \centering
  \caption{The classification performance with and without the marginal loss (\%)}
    \begin{tabular}{cccccccc}
    \toprule
    Methods & Mutag & Mutagen & PTC-FM & PTC-FR & PTC-MM & PTC-MR & Liver \\
    \midrule
    \emph{Lamole}-$\mathcal{L}_{M}$ & 84.9  & 73.8  & 58.6  & 69.3  & 59.7  & 62.3  & 44.6 \\
    \emph{Lamole} & \textbf{88.2} & \textbf{74.5} & \textbf{62.4} & \textbf{70.0} & \textbf{61.2} & \textbf{66.0} & \textbf{47.5} \\
    \bottomrule
    \end{tabular}%
  \label{tab:woLMclass}%
\end{table}%

To investigate the attention weights, the attention weights of two molecules are depicted in Fig. \ref{fig:attenvis}. From Fig. \ref{fig:attenvis} right panel, it can be found that the correlation among the ground truth substructures is higher than others, showcasing the rationality of using attention weights to capture the functional group interactions. When we aggregate the attention weights for each token, the top two attention weights of the molecule (ID:155) match the two ground truth substructures (see Fig. \ref{fig:attenvis} (a) left bottom panel). However, the top two attention weights of the molecule (ID:156) do not match the two ground truth substructures (see Fig. \ref{fig:attenvis} (b) left bottom panel). The above results indicate that attention weights can capture the interactions and also confirm that "attention is not explanation"~\cite{jain2019attention,serrano2019attention,abnar2020quantifying}. 

\begin{figure}[t] 
  \centering   
  \includegraphics[width=\linewidth]{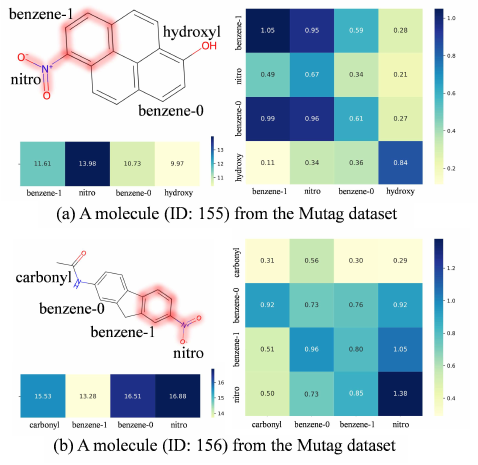}
  \caption{Attention weights visualization. The ground truth substructures are highlighted in red.}

  \label{fig:attenvis}
\end{figure}

 \begin{figure*}[!t] 
  \centering   
  \includegraphics[width=14.2cm]{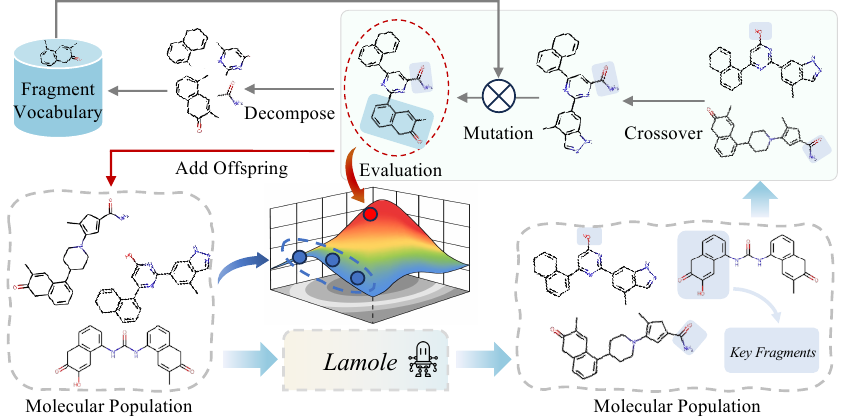}
  \caption{The proposed interpretable molecular optimization pipeline integrates {\em Lamole}'s explainable molecular property prediction with an evolutionary algorithm to guide the editing process: The process begins with {\em Lamole} identifying the key functional groups in seed molecules. In each generation, crossover swaps the key fragments between parents, while mutation introduces novel fragments either randomly or by using a fragment vocabulary. The selection process retains the promising molecules for the next generation. In this loop, {\em Lamole} identifies new key groups and dynamically updates the fragment vocabulary for guiding the optimization.}
  \label{fig:EAediting}
\end{figure*}

\subsection{Computational Cost}
The pre-training stage took 11.8 hours. After pre-training, we fine-tuned the model on the used dataset. On the Mutag dataset, this process took 158s, and the evaluation time was 15s. For the baseline methods, SmoothGrad+GCN and OrphicX, the total training and evaluation time is 87s and 122s, respectively. Considering that the language model only requires pretraining once, the proposed method consumes an acceptable additional computational cost but brings comparable classification accuracy and explainability.

\subsection{Case Study: Actionable Utility for Molecular Editing}

The above experiments demonstrate that our proposed {\em Lamole} could provide chemical concepts-aligned explanations. This feature is crucial, particularly for practical use like molecular editing, as a concept-aligned explanation is inherently actionable for humans --- these explanations can directly guide the modification of molecules. In this section, we illustrate the actionable utility of {\em Lamole}'s explanations in molecular editing tasks. Specifically, this study focuses on optimizing ADME properties. We achieve this by establishing an interpretable molecular optimization pipeline that integrates {\em Lamole}'s explainable molecular property prediction with an evolutionary algorithm to guide the editing process.

The interpretable molecular optimization pipeline begins by defining the optimization objective. This objective corresponds to the property score, $F(\cdot)$, which serves as the fitness function for the evolutionary algorithm. For this study, we chose solubility and permeability as the optimization objectives. For instance, maximizing the solubility score or the permeability score. 
As presented in Fig. \ref{fig:EAediting}, firstly, we initialize the evolutionary algorithm by selecting $N=20$ seed molecules from the ZINC dataset to form the initial population, $P_0$. For each molecule $M \in P_0$, the {\em Lamole} framework is immediately applied to identify the most influential functional groups that contribute to the property score, thereby informing the subsequent genetic operators.

In each generation $t$, a new population $P_{t+1}$ is derived from the current population $P_t$ through the fragment-informed genetic operators. These operators leverage the key fragments provided by {\em Lamole} to bias the search towards chemically meaningful modifications, ensuring the evolution is both effective and interpretable.

\begin{itemize}
    \item Crossover: the crossover operator is designed to facilitate the targeted exchange of critical structural information. Two parent molecules, $M_1$ and $M_2$, are selected from $P_t$. Instead of performing random recombination, the operator utilizes {\em Lamole} to identify the key functional groups contributing to the target property. These highly influential substructures are then swapped between $M_1$ and $M_2$ to generate two novel offspring, $M_1'$ and $M_2'$. This mechanism ensures that the recombination step targets the most predictive parts of the molecular structure, thereby enhancing the probability of generating molecules with desired properties in the subsequent generation.
    
    \item The mutation operator introduces structural diversity to explore the chemical landscape. We employ two distinct mutation strategies: 1) Random structural mutation: This mode ensures broad exploration. A molecule $M \in P_t$ is selected, and one of its constituent fragments encoded via Group SELFIES is arbitrarily replaced. This is essential for preventing the algorithm from prematurely converging to local optima. 2) Vocabulary-based mutation: This strategy is designed to enhance the chemical plausibility and efficiency of the search by leveraging historical knowledge. We first define the fragment vocabulary ($V$) as the set of chemical fragments that have been identified as key contributors by {\em Lamole}. The vocabulary is maintained and updated throughout the evolutionary history. During mutation, a molecule $M$ is selected, and one of its existing fragments is substituted with a fragment retrieved from this vocabulary $V$. The merits of retrieving fragments from the vocabulary $V$ are twofold: (1) it ensures that the introduced structural change is biased towards known chemical building blocks that have been validated during the optimization process; (2) it balances exploration (introducing new fragments) with exploitation (recombining chemically validated motifs).
\end{itemize}

Following the execution of crossover and mutation operators, a large pool of candidate offspring molecules is generated. These generated offspring are then subjected to property score evaluation. The target property score, $F(M')$, for each new molecule $M' \in P_{t+1}$ is calculated using $\text{RDKit}$\footnote{\textcolor{black}{Because we use the RDKit-computed objective, this molecular editing case is presented only as a computational demonstration, and the resulting optimized molecules do not imply real-world chemical optimality.}}. Simultaneously, the {\em Lamole} model is executed on these new molecules to identify the emergent key fragments that contribute to their scores. If a new molecule exhibits a higher property score than its parents and contains a fragment not yet present in the vocabulary ($V$), then the fragment is updated into $V$ for use in future mutation steps. Finally, based on the property score $F(M)$, a selection scheme is applied to $P_t \cup P_{t+1}$ to determine which molecules form the population $P_{t+1}$ for the next generation. This entire iterative process continues until a predefined maximum number of generations $t_{max}=50$ is reached.

To benchmark the efficacy of the proposed interpretable molecular editing pipeline, we compare our pipeline against several baselines, and all baselines are from \cite{liu2023multi}. 1) Random: This baseline applies a random noise as the perturbation to the latent representation, testing optimization performance achieved through purely random exploration in the latent space. 2) PCA (Principal Component Analysis): This method guides modifications using the eigenvectors obtained from the $\text{PCA}$ decomposition of the latent representation. This approach tests optimization guided by the directions of maximum global variance in the representation space. 3) High-Variance: This technique attempts to target specific semantic directions by focusing on the latent representation dimension with the highest variance. 4) MoleculeSTM: This method utilizes a multimodal structure–text model by jointly learning molecules’ chemical structures and textual descriptions via a contrastive learning strategy. 5) Genetic Search: It directly applies genetic operators to the molecular graph representation.

\begin{table*}[htbp]
  \centering
  \caption{Results on four molecule editing tasks, and the evaluation is the hit ratio of the property change.}
    \begin{tabular}{ccccccc}
    \toprule
    Target Property & Random & PCA   & High Variance & GS-Mutate & MoleculeSTM & Ours \\
    \midrule
    Higher Permeability & 25.23 & 21.36 & 21.98 & 22.00 & 31.76 & \textbf{39.55} \\
    Lower Permeability & 16.79 & 15.48 & 14.66 & 28.83 & 37.35 & \textbf{45.90} \\
    More Soluble in Water & 35.33 & 33.80 & 33.52 & 52.00 & 67.86 & \textbf{70.83} \\
    Less Soluble in Water & 43.36 & 39.36 & 42.89 & 47.50 & 64.79 & \textbf{66.33} \\
    \bottomrule
    \end{tabular}%
  \label{tab:editing}%
\end{table*}%

\begin{figure*}[htbp] 
  \centering   
  \includegraphics[width=16.8cm]{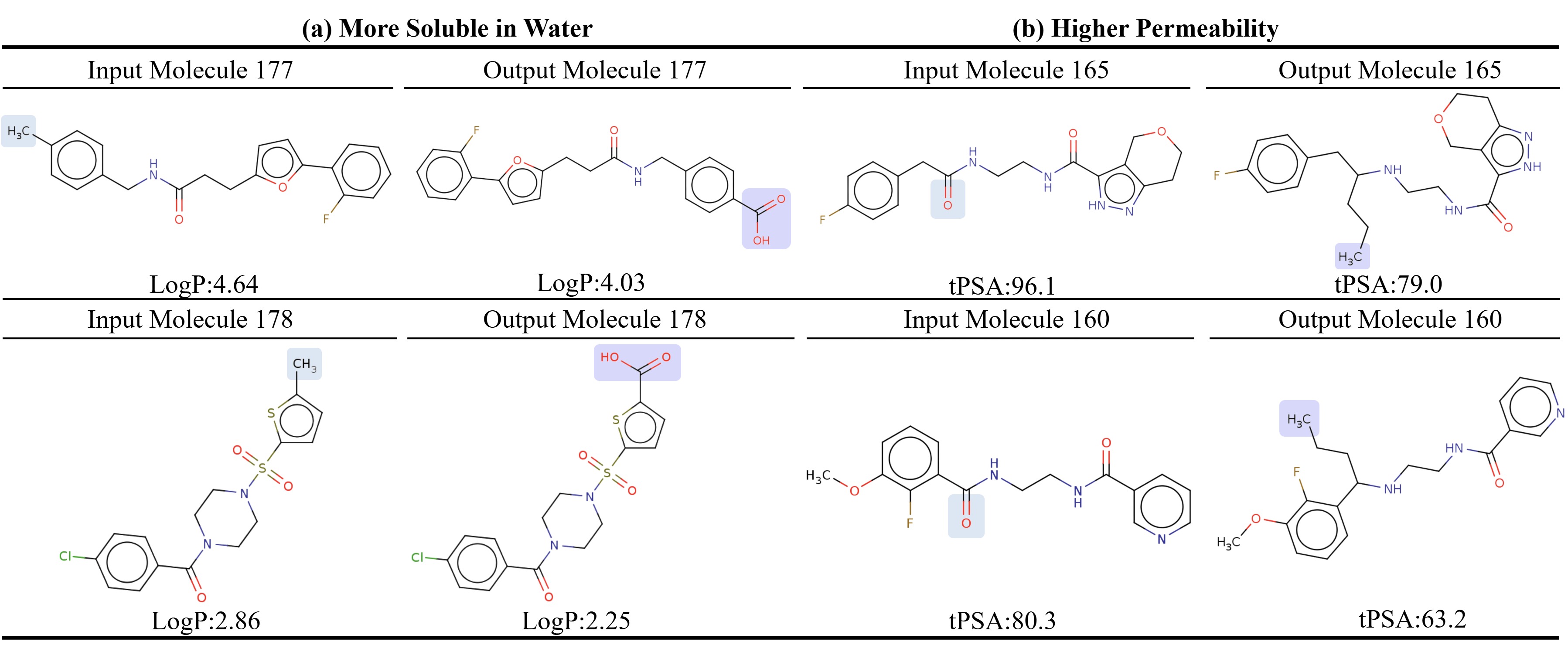}
  \caption{Visualization of the molecule editing tasks. The blue region corresponds to the edited substructures in the molecule.}
  \label{fig:edit_up}
\end{figure*}

The results on the molecular editing task are summarized in Table \ref{tab:editing}, which reports the hit ratio of achieving the target property change for four distinct optimization goals. Across all four objectives—increasing or decreasing permeability and increasing or decreasing water solubility—our {\em Lamole}-guided evolutionary algorithm achieved the highest hit ratios. For enhancing permeability, our method yielded a hit ratio of $39.55\%$, surpassing the best baseline, $\text{MoleculeSTM}$. The most substantial gain was observed when optimizing for lower permeability, where our method reached $45.90\%$, higher than $\text{MoleculeSTM}$ ($37.35\%$) and $\text{GS-Mutate}$ ($28.83\%$). Similarly, for solubility, our approach demonstrated superior efficiency. For achieving more soluble in water, the hit ratio was $70.83\%$, narrowly exceeding $\text{MoleculeSTM}$ and showing a wide margin over $\text{GS-Mutate}$. When optimizing for less soluble in water, our method maintained the lead with a hit ratio of $66.33\%$. Fig. \ref{fig:edit_up} presents four representative examples of molecular editing, illustrating the resulting structures optimized for higher solubility in water and higher permeability, respectively. In both cases, the properties show a notable increase following the targeted fragment modifications. The corresponding visualizations for edits targeting lower solubility in water and lower permeability are provided in the supplementary material. The performance across all objectives demonstrates that utilizing the chemically informed key fragments identified by {\em Lamole} effectively guides the evolutionary search.

\section{Conclusions}

In this work, we presented {\em Lamole}, a novel framework for explainable molecular property prediction based on LMs, specifically designed to address the critical gaps in generating chemically meaningful and faithful explanations. To overcome the representational limitations of existing methods, we first leveraged the Group SELFIES string representation, which intrinsically encodes molecular information at the functional group level, aligning model input directly with chemical intuition. Furthermore, we analyzed the deficiencies of relying solely on attention or gradients and proposed an information flow-based explanation method by carefully disentangling the internal architecture of the Transformer. This unique approach integrates both self-attention weights and gradients to accurately quantify the contribution of each functional substructure and capture inter-group interactions. To ensure the plausibility of the explanation, we introduced a novel marginal loss designed to calibrate the explanations against expert annotations. We theoretically demonstrated that this loss aligns the generated explanations with the tangent space of the data manifold, ensuring concept-aligned and robust results. 

Empirical validation across eight diverse datasets, including toxicity, solubility, and permeability, confirmed that {\em Lamole} achieves predictive accuracy comparable to state-of-the-art baselines while significantly boosting explanation accuracy. We also showcased the practical, actionable utility of our framework by integrating {\em Lamole} with an evolutionary algorithm, establishing an interpretable optimization pipeline that serves as a rational guide for molecular editing and design. \textcolor{black}{This work establishes a concept-aligned explainability framework on standard molecular benchmarks. However, extending it to broader generalization scenarios, including unseen chemotypes, remains an important direction for future work.}

\bibliography{example_paper}
\bibliographystyle{IEEEtran}

\end{document}